\documentclass[fleqn,10pt]{wlscirep}
\usepackage{indentfirst}
\usepackage{amsmath}  
\usepackage{amssymb}  
\usepackage{newunicodechar}
\usepackage{array} 
\usepackage{amsmath}
\usepackage{amsfonts}
\usepackage{multirow}
\usepackage{booktabs}
\usepackage{graphicx}
\usepackage[utf8]{inputenc}
\usepackage[T1]{fontenc}

\usepackage{enumitem}
\usepackage{cite}
\usepackage{geometry}
\usepackage{float} 
\usepackage{etoolbox}
\usepackage{newunicodechar}
\newunicodechar{：}{:}
\usepackage{multirow}    
\usepackage{authblk}

\title{FuXi-Air: Urban Air Quality Forecasting Based on Emission-Meteorology-Pollutant multimodal Machine Learning}

\author[1,*]{Zhixin Geng}
\author[2,*]{Xu Fan}
\author[1]{Xiqiao Lu}
\author[1,5,6,+]{Yan Zhang}
\author[3]{Guangyuan Yu}
\author[3,+]{Yuewu Li}
\author[3]{Qian Wang}
\author[3]{Cheng Huang}
\author[1]{Weichun Ma}
\author[1]{Qi Yu}
\author[5,2]{Libo Wu}
\author[4,2,+]{Hao Li}

\affil[1]{Shanghai Key Laboratory of Atmospheric Particle Pollution and Prevention, National Observations and Research Station for Wetland Ecosystems of the Yangtze Estuary, Department of Environmental Science and Engineering, Fudan University, Shanghai 200433, China}
\affil[2]{Shanghai Academy of AI for Science (SAIS), Shanghai 200003, China}
\affil[3]{Shanghai Environmental Monitoring Center (SEMC), Shanghai 200235, China}
\affil[4]{Artificial Intelligence Innovation and Incubation (AI3) Institute of Fudan University, Shanghai 200433, China}
\affil[5]{MOE Laboratory for National Development and Intelligent Governance, Shanghai Institute for Energy and Carbon Neutrality Strategy, IRDR ICoE on Risk Interconnectivity and Governance on Weather/Climate Extremes Impact and Public Health, Fudan University, Shanghai 200433, China}
\affil[6]{Shanghai Institute of Eco-Chongming (SIEC), Shanghai, 200062, China}
\affil[+]{Corresponding to：yan\_zhang@fudan.edu.cn; lihao\_lh@fudan.edu.cn; yuewuli5@163.com}
\affil[*]{These authors contributed equally to this work.}

\begin{abstract}
Air pollution has emerged as a major public health challenge in megacities. Numerical simulations and single-site machine learning approaches have been widely applied in air quality forecasting tasks. However, these methods face multiple limitations, including high computational costs, low operational efficiency, and limited integration with observational data. With the rapid advancement of artificial intelligence, there is an urgent need to develop a low-cost, efficient air quality forecasting model composed of a smart urban management system. An air quality forecasting model, named FuXi-Air, has been constructed in this study based on multimodal data fusion to support highly precision and air quality forecasting and operated in typical megacities. The model integrates meteorological forecasts, emission inventories, and pollutant monitoring data under the guidance of air pollution mechanism. By combining an autoregressive prediction framework with a frame interpolation strategy, the model successfully completes 72-hour forecasts for six major air pollutants at an hourly resolution across multiple monitoring sites within 25–30 seconds. In terms of both computational efficiency and forecasting accuracy, it outperforms the mainstream numerical air quality models in operational forecasting work. Ablation experiments concerning the key influencing factors show that although meteorological data contribute more to model accuracy than emission inventories do, the integration of multimodal data significantly improves the precision of forecasting and ensures that reliable predictions are obtained under differing pollution mechanisms across megacities. This study provides both a technical reference and a practical example for applying multimodal data-driven models to air quality forecasting and offers new insights into building hybrid forecasting systems to support air pollution risk warning for a smart city management.
\end{abstract}

\begin{document}
\maketitle
\newpage

\section*{Highlights}
\begin{itemize}
 \item A novel multimodal data-driven air quality model achieved highly spatiotemporally accurate operational forecasting in megacities.
 \item A transformer network presented the complex interactions among meteorology, emissions and air pollutant observations.
 \item Performing flexible offline integration with AI-based meteorological models extended the potential forecasting horizon.
 
\end{itemize}

\section{Introduction}

Air pollution has emerged as a major global challenge with profound implications for public health, ecosystem stability, and socioeconomic development. In urban environments, where human activities are densely concentrated, air pollution represents an especially critical issue that threatens the health and well-being of millions of city dwellers. Key pollutants, including fine particulate matter (PM$_{2.5}$), ozone (O$_{3}$), and nitrogen dioxide (NO$_{2}$), pose severe risks to the human respiratory and cardiovascular systems of humans while also accelerating climate change and disturbing the ecological balance \cite{1chen2021possible,2xu2025rising,3feng2024long,4epa2019isa}. Urban air pollution exacerbates the burden on city infrastructure, contributes to health disparities among socioeconomically disadvantaged populations, and hampers sustainable urban development. The acceleration of urbanization and increasing industrial emissions have contributed to more frequent episodes of severe air pollution, increasing the demand for accurate forecasting systems\cite{5liang2019urbanization,6wang2021anthropogenic}. High-quality air pollution forecasts provide essential scientific policy-making support, enabling authorities to formulate emission reduction strategies, optimize pollution control measures, and safeguard public health\cite{7amann2011cost,8ekeh2025leveraging,9yang2023novel,10ding2022forecasting}.

Meteorological conditions play decisive roles in the formation, accumulation, and dispersion of air pollutants\cite{11liu2020exploring,12wen2024assessing}. Urban meteorology is often characterized by complex boundary layer dynamics, heat island effects, and localized turbulence caused by high building density, all of which complicate pollutant transport and mixing processes. The speed and direction of wind affect the horizontal transport of pollutants\cite{13mai2024convolutional,14zhou2024impacts}, temperature and humidity influence their chemical transformations\cite{15han2020local,16schnell2018exploring}; and the height of the planetary boundary layer determines the extent of vertical mixing\cite{17yuval2020association,18miao2021relationship}.The nonlinear coupling between meteorological drivers and pollutant dynamics is critical for attaining improved forecasting accuracy. The current air quality forecasting systems are based predominantly on numerical models rooted in physicochemical equations, which describe pollutant formation, transport, and removal processes in considerable detail\cite{19zhang2023evolution,20sulaymon2023using,21tie2010impact}. However, these models suffer from high computational costs and relatively low efficiency. Their predictive performance is often limited by the accuracy of their meteorological inputs and by weak integration with real-time observational data, which impairs their operational scalability and responsiveness in fast-changing urban environments\cite{22wang2025causal,23li2019ozone,24zhang2020source,25yu2010eta}. In recent years, deep learning-based weather forecasting models, such as FuXi\cite{27chen2023fuxi}, GraphCast\cite{28lam2023learning}, and FourCastNet\cite{29pathak2022fourcastnet} , have outperformed the traditional numerical forecasting methods in terms of precision. These models require fewer computational resources and offer higher efficiency than that of their predecessors while providing regionally gridded, high-resolution meteorological forecasts. This high spatial resolution is particularly well-suited for urban areas, where pollutant concentrations can vary significantly over short distances. Such capabilities enable real-time coupling with air quality models at the urban scale, making urban-scale air pollutant forecasting more feasible.

With the continuous advancement of machine learning technologies, machine learning–based methods have been increasingly applied to pollutant concentration forecasting and source apportionment. The current research is focused primarily on short-term prediction models involving either multisite, single-pollutant or single-site, single-pollutant scenarios, along with the identification and analysis of the key driving factors. Although notable progress has been made in terms of model architecture development and predictive performance enhancement, several limitations remain, including generalizability limitations, significant error accumulation, and the reliance on a narrow range of auxiliary data. Recent studies have integrated advanced deep learning architectures to address the various challenges encountered in air quality forecasting tasks. Wu et al.\cite{wu2023hybrid} combined ResNet, graph convolutional networks (GCNs) and bidirectional long short-term memory (BiLSTM) networks to capture temporal dependencies, spatial topologies, and meteorological influences, respectively. However, the long-range multistep forecasting ability of this model was not validated, and it tended to underestimate extreme O$_{3}$ levels. Cheng et al.\cite{cheng2021development} integrated wavelet decomposition, gated recurrent units (GRUs) and support vector regression (SVR) into the WD-GRU-SVR model and demonstrated the applicability of this model across six representative Chinese cities. Despite its innovation, this model was focused solely on O$_{3}$ and failed to consider interpollutant interactions, with performance degradations observed in coastal cities. In a separate study, the GC-LSTM model, which combines GCNs and long short-term memory (LSTM) networks, was developed to represent the spatiotemporal information across monitoring stations as graph signals\cite{qi2019hybrid}. While achieving a forecasting horizon of up to 72 hours, this model experienced substantial error accumulation in long-range prediction scenarios. Microsoft developed the Aurora global atmospheric composition forecasting system\cite{bodnar2025foundation}, providing a reference architecture for global-scale air quality forecasting. Despite its use of advanced deep learning frameworks and its strong performance across various atmospheric constituents, the model does not incorporate emission inventory data or explicitly account for the physical and chemical mechanisms underlying the pollutant formation and transformation processes. Importantly, atmospheric pollutants exhibit short lifespans, strong urban variability, and high sensitivity to local meteorological conditions. These properties give rise to nonstationary and heterogeneous spatiotemporal pollutant distributions. These challenges are particularly pronounced in urban contexts, where emissions are highly localized and meteorological conditions can vary drastically over short distances. Accordingly, developing high-resolution forecasting models that incorporate mechanistic insights—especially at the urban scale—is essential for improving both the scientific credibility and practical utility of air quality prediction systems.

Among the emerging machine learning techniques, the self-attention mechanism and graph neural networks (GNNs) have become two dominant modeling paradigms for pollutant concentration forecasting\cite{30zhu2024hybrid,31mandal2023city,32zhang2024long,33liu2022prediction,34guyu2025pm2}. GNNs typically rely on prior knowledge to construct spatial relationship graphs between monitoring sites, thereby simulating the transport dynamics of pollutants. However, the effectiveness of GNNs is often constrained by urban heterogeneity, pollutant-specific behaviors, and dynamically changing meteorological conditions, which limits their adaptability in terms of capturing cross-urban pollutant correlations\cite{35zhang2023air}. In contrast, the self-attention mechanism can learn the spatiotemporal dependencies between sites in a purely data-driven manner, offering greater flexibility when modeling the transport of pollutants such as PM$_{2.5}$ and PM$_{10}$ \cite{36wang2022air}. Nonetheless, the current deep learning models continue to face several challenges. These include limited multipollutant joint modeling capacities, the accumulation of significant errors in long-sequence predictions, the weak integration of meteorological-emission coupling mechanisms, and poor generalizability across different cities. In urban settings, the ability to jointly model pollutants and incorporate dynamic emission sources, such as vehicular and industrial emissions, is essential for accurate and timely air quality forecasting. On the one hand, most models rely heavily on historical pollutant data while failing to effectively incorporate high-resolution meteorological forecasts and dynamic emission inventories. This limits their ability to reflect the underlying processes of pollution formation. On the other hand, variations in the pollution sources and meteorological conditions observed across cities lead to model performance degradations when applied outside their training domains. As a result, these models struggle to capture nonlinear inter-pollutants interactions—such as the photochemical reactions between NO$_{2}$ and  O$_{3}$—and are prone to cumulative forecasting errors, leading to inaccurate pollution intensity predictions and delayed early warnings\cite{37baruah2024novel,38li2024multi,39liu2024air}. Therefore, the development of forecasting methods that are accurate, efficient, and adaptive across cities has become a pressing research priority at the intersection of environmental science and artificial intelligence.

In summary, a meteorology-emission coupled multitemporal-scale air quality forecasting model (FuXi-Air), which integrates multimodal data fusion with a self-attention mechanism, is proposed in this study. FuXi meteorological forecast data, dynamic emission inventories (CAMS), and pollutant monitoring observations are coupled within a unified framework. The emission inventory provides the spatial distributions of pollutant emissions, whereas the attention mechanism dynamically captures the influences of meteorological conditions on pollutant transport and chemical transformation, thereby enhancing the capacity of the model to represent complex pollution processes. By combining an autoregressive forecasting module with a frame interpolation strategy, the model reduces its reliance on input data and enables the joint forecasting of multiple pollutants with an hourly resolution across multiple monitoring sites. A quantile loss function is further introduced to improve the predictive performance of the model, allowing the model to capture high-pollution risks at different quantiles and to quantify forecast uncertainty. Additionally, to address the unique air pollution challenges faced by cities, a series of experiments are designed to validate the critical roles of meteorological factors and emission inventories in forecasting key pollutants at urban monitoring sites. The results confirm the advantages of joint multipollutant prediction and demonstrate the generalizability of the FuXi-Air model across different cities, as well as its ability to reflect variations in the underlying pollution mechanisms. This research provides theoretical support for integrating mechanism-driven and data-driven approaches to develop high-precision, generalizable AI-based air pollutant forecasting systems and offers a technical pathway for advancing deep learning–based data assimilation frameworks in the field of urban-scale air quality prediction.

\section{Data and Methods}

\subsection{Forecasting Techniques and Methods}

\begin{figure}[H]
    \centering
    \includegraphics[width=\linewidth]{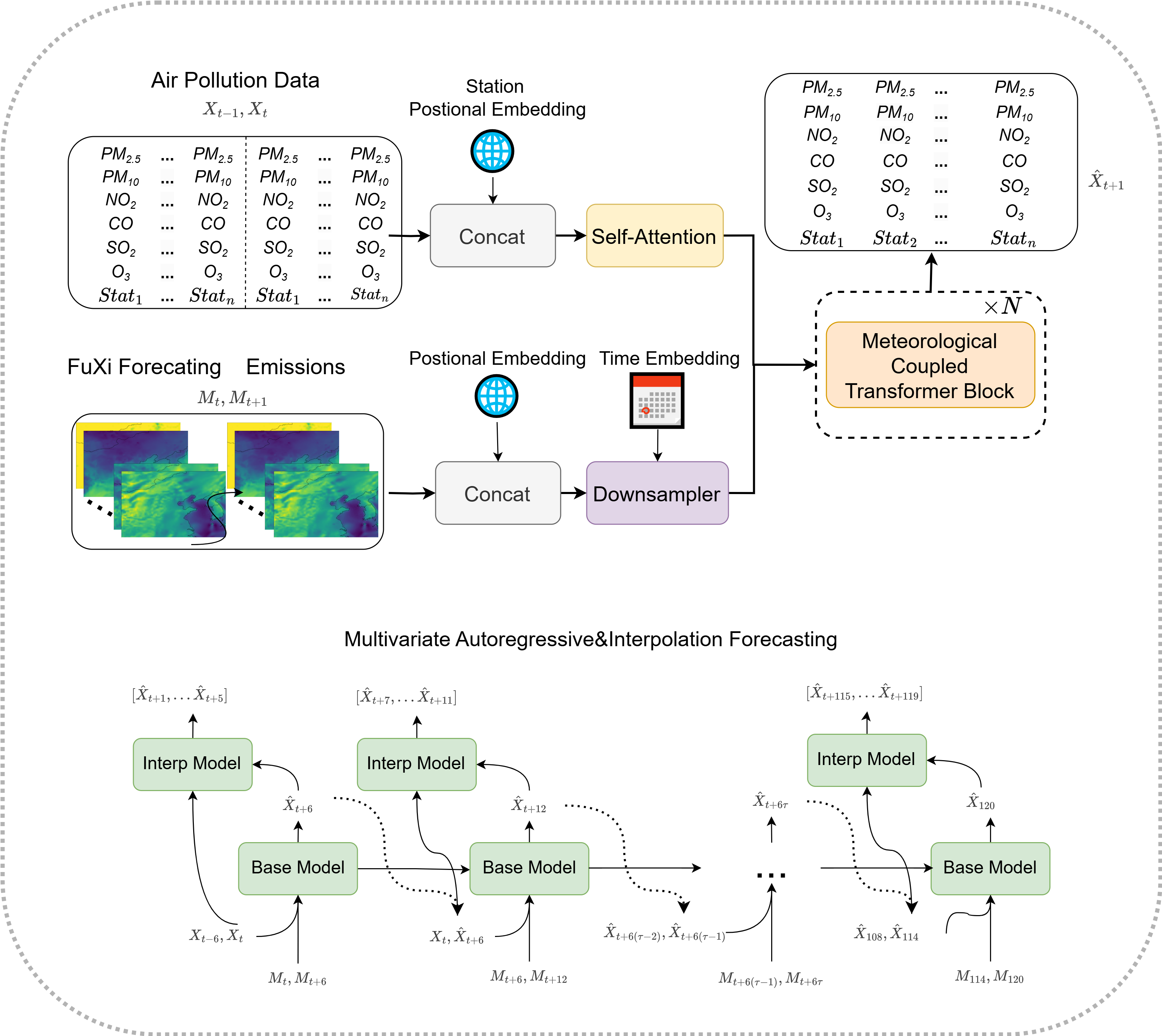}
    \caption{Schematic diagram of the proposed model structure.}
    \label{fig 1:model_structure}
\end{figure}

A pollutant–emission–meteorology coupling module is designed in this study based on an attention mechanism; for the first time, site-level pollutant forecasting is integrated with AI-based meteorological prediction. The module fully leverages the high-resolution meteorological fields provided by the FuXi model and explicitly considers the interactions between meteorological conditions and pollutant emissions, thus enhancing the ability of the model to represent complex environmental factors. The resulting framework is referred to as the A meteorology–emission–pollutant coupled air quality forecasting model (FuXi-Air). Considering the periodic variation characteristics of atmospheric pollutants, a hybrid forecasting framework is constructed in this study, and its core consists of the following: an autoregressive prediction model with a 6-hour resolution serving as the primary forecasting module for capturing long-term trends and a frame interpolation model that further refines the 6-hour predictions to achieve pollutant forecasting at an hourly resolution(Fig. ~\ref{fig 1:model_structure}).

\subsection{Study Area and Selected Cities}

\begin{figure}[H]
    \centering
    \includegraphics[width=\linewidth]{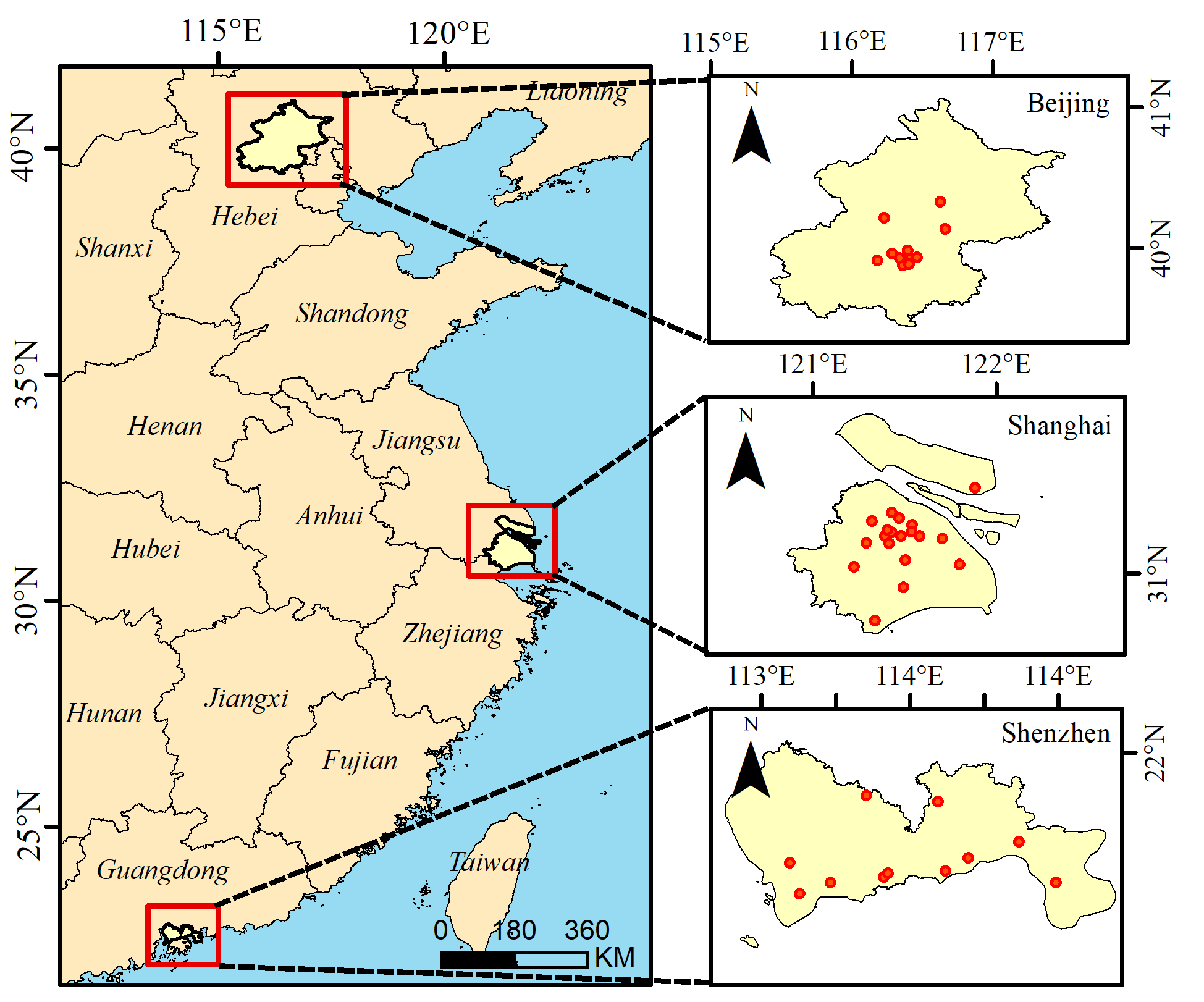}
    \caption{Study area and site distribution map.}
    \label{fig:2.2study_area}
\end{figure}

Three representative cities in China—Beijing, Shanghai, and Shenzhen—are selected as the target cities in this study (Fig. ~\ref{fig:2.2study_area}). These cities differ significantly in terms of their geographic locations, climate types, pollutant compositions, and emission source profiles, offering strong regional diversity and representativeness. Beijing, which is located inland in northern China, has a temperate monsoon climate and is known as a typical heavily polluted city\cite{zhang2016air}. Its air quality is strongly affected by wintertime heating emissions and stagnant meteorological conditions, resulting in frequent and severe episodes of particulate pollution (PM$_{2.5}$ and PM$_{10}$)\cite{feng2014formation}. Shanghai, which is situated in the Yangtze River estuary, experiences a subtropical monsoon climate. It is characterized by pronounced O$_{3}$ pollution during Summer, with its pollutant levels showing strong sensitivity to meteorological conditions such as temperature and solar radiation\cite{chang2021meteorology}. Shenzhen is located on the southern coastal margin and features a subtropical marine climate with relatively low overall pollution levels, high atmospheric ventilation, and low pollutant concentration variability. Among these three cities, Beijing represents a typical inland city, whereas Shanghai and Shenzhen are coastal cities. Together, they constitute a set of diverse pollution scenarios, which is highly important for evaluating the generalization performance of the proposed model under different regional pollution backgrounds.

\subsection{Dataset Construction}

Three main data sources are utilized in this study for the cities of Beijing, Shanghai, and Shenzhen: (1)\quad national air quality monitoring station data concerning pollutant concentrations, (2)\quad atmospheric pollutant monitoring and emission inventory data from the Copernicus Atmosphere Monitoring Service (CAMS), and (3)\quad meteorological forecast data from the FuXi-2.0 model. The dataset spans the period from 2016 to 2023, with the data from 2016 to 2022 used for training and the data from 2023 used for testing. The national air quality monitoring dataset includes hourly concentration measurements of six pollutants: SO$_{2}$, NO$_{2}$, CO, O$_{3}$, PM$_{2.5}$, and PM$_{10}$. The selected CAMS emission inventory includes seven variables that are closely related to the target pollutants: NO$_x$, CO, NH$_3$, PM$_{10}$, PM$_{2.5}$, SO$_2$, and VOCs. The gridded CAMS inventory is interpolated to a spatial resolution of 0.1° × 0.1°, with a temporal resolution of one month. Twelve meteorological variables derived from the FuXi-2.0 model outputs at pressure levels of 1000hPa, 925hPa, and 850hPa are selected. The gridded meteorological data obtained from FuXi are interpolated to a spatial resolution of 0.1° × 0.1° and have a temporal resolution of 1 hour. To address the missing values in the pollutant concentration monitoring data, a data preprocessing scheme is proposed. More detailed descriptions of the data can be found in the Supplementary materials A. To improve the predictive performance of the developed model, the concentrations of the six pollutants are standardized using the following formula:

\begin{equation}
x^{\text{norm}}_{i,t} = \frac{x_{i,t} - \text{mean}(\tilde{X}_i)}{\text{std}(\tilde{X}_i)}
\end{equation}

where “mean” and STD denote the mean and standard deviation computed along the temporal dimension, respectively, and $\tilde{X}_i$ represents the full historical dataset for city i.

\subsection{Simulation and Forecast Evaluation Methods}

In this study, data from 2023 are used to comprehensively evaluate the forecasting performance achieved by the model for six pollutants across three major cities. Five commonly used statistical indicators are selected for assessment purposes.

The correlation coefficient (R) is used to evaluate the fitting ability of the model; the root mean square error (RMSE) is used to analyze the actual prediction accuracy of the model, and the relative root mean square error (rRMSE) is used to normalize the RMSE by the mean of observations, allowing for a unitless comparison across different pollutants or cities. The mean absolute error (MAE) is used to assess the overall error level, and the mean relative error (MRE) is used to compare the relative differences between cities.

\section{Results and Discussion}

\subsection{Design and Construction of the Pollutant Forecasting Model}

\subsubsection{Dual-Scale Air Quality Forecasting Methods}  

Air pollutant monitoring data often contain substantial missing values and outliers. The direct application of strict quality control and imputation strategies may introduce additional errors, thereby compromising the actual predictive accuracy of the model. Therefore, reducing the dependence of the model on the input data represents a more robust solution, which helps mitigate the impact of data quality issues on forecasting performance and improves the stability and reliability of the model in operational settings.

To minimize its reliance on the input data, the base forecasting model uses only two temporal inputs:$X_{t-6}$ and $X_{t}$. An autoregressive strategy is employed to generate 6-hour-resolution pollutant concentration forecasts over a 72-hour period (12 prediction steps). On this basis, a frame interpolation model is further employed to refine the outputs of the 6-hour model by infilling the intermediate five hours between each pair of prediction steps, thereby enhancing the accuracy of the hourly forecasts. The architecture of the interpolation model is consistent with that of the base 6-hour forecasting model, but it uses different supervisory signals to accommodate the requirements of forecasting at different temporal scales.

\subsubsection{Pollutant Monitoring Site Interaction Module} 

To enhance the ability of the model to learn the spatial relationships various among monitoring sites and to accurately characterize the inter-pollutant differences within spatial responses, a site interaction module is constructed in this study based on a self-attention mechanism. This module is designed to extract more representative latent features and to optimize the coupling of meteorological and emission inventory data. Specifically, the pollutant concentration data acquired from two-time steps are concatenated along the temporal dimension. These features are then combined with relative positional encodings based on the latitude and longitude differences between sites, and the resulting representation is input into a self-attention layer to capture the spatial correlations among the monitoring sites. In addition, considering that atmospheric pollutant variations are strongly influenced by both diurnal and seasonal cycles, we incorporate temporal encodings into the input features. The time variable t is embedded using an embedding method that includes two key components: the day of the year (DOY) and the hour of the day (HOD). This enables the model to better capture temporal variation patterns and thereby improve its predictive performance. The interaction features generated by the self-attention module are subsequently used as inputs for the meteorology–emission coupling module, further enhancing the capacity of the model to learn the spatiotemporal evolution trends of pollutant concentrations.

\subsubsection{Meteorology–Emission Inventory Coupling Module}

Meteorological fields influence the behaviors of pollutant emission inventories by modulating the transport effects of pollutants. Different meteorological conditions alter the transport mechanisms of pollutants between monitoring sites, as well as the interaction patterns among different pollutant species. Therefore, a module for coupling pollutant site-level latent variables and meteorological–emission fields is developed in this study based on an attention mechanism to dynamically capture the impacts of meteorological conditions on the pollutant transport and transformation processes. Specifically, the module first employs a convolutional neural network (CNN) architecture based on a residual network (ResNet) to extract features from the gridded meteorological forecasts provided by the FuXi meteorology AI model and from the emission inventory data, thereby obtaining multiscale spatiotemporal feature representations. On this basis, a cross-attention mechanism is designed, where the latent interaction variables obtained from the site module (Hsa, see Supplementary materials B) are used as query to extract informative features from the latent meteorological and emission variables. In the residual connection component, the query is refined by adding it to the features extracted via the cross-attention mechanism. This design enhances the representational capacity of the model while preserving the original input information.

\subsubsection{Model Training and Evaluation Processes}

In air pollutant forecasting cases, pollutant concentration uncertainty remains a critical challenge. Due to the influences of various complex factors—including meteorological conditions, emission source variations, chemical reactions, and topographical effects—a single deterministic forecast often fails to accurately represent actual pollution levels. This is especially problematic during extreme pollution events (e.g., heavy or explosive pollution episodes), where such forecasts may underestimate risks or delay warnings, thereby impairing the effectiveness of environmental management decisions. As a result, quantifying forecast uncertainty is essential for operational air quality prediction tasks. Quantile forecasting offers an effective solution. It not only captures the central tendencies of pollutant concentrations (e.g., the 50th percentile) but also enables the identification of high-risk scenarios by predicting values at different quantiles (such as the 10th or 90th percentile), thereby allowing uncertainty to be explicitly quantified. To effectively learn and optimize predictions at different quantiles, a quantile loss function is introduced in this study. Compared with traditional metrics such as the mean squared error (MSE) or mean absolute error (MAE), the quantile loss allows for an asymmetric optimization process that is tailored to varying pollution levels. This enables the model to learn the distributional characteristics of pollutant concentrations across different quantiles, thereby improving its sensitivity to extreme pollution events and enhancing the overall stability and reliability of the output forecasts.

\subsection{Evaluation of the Applicability and Performance of the FuXi-Air Model in Different Megacities of China}

\subsubsection{Evaluation of the Predictive Performance of the Model for Six Air Pollutants}  

\begin{figure}[H]
\centering
\includegraphics[width=\linewidth]{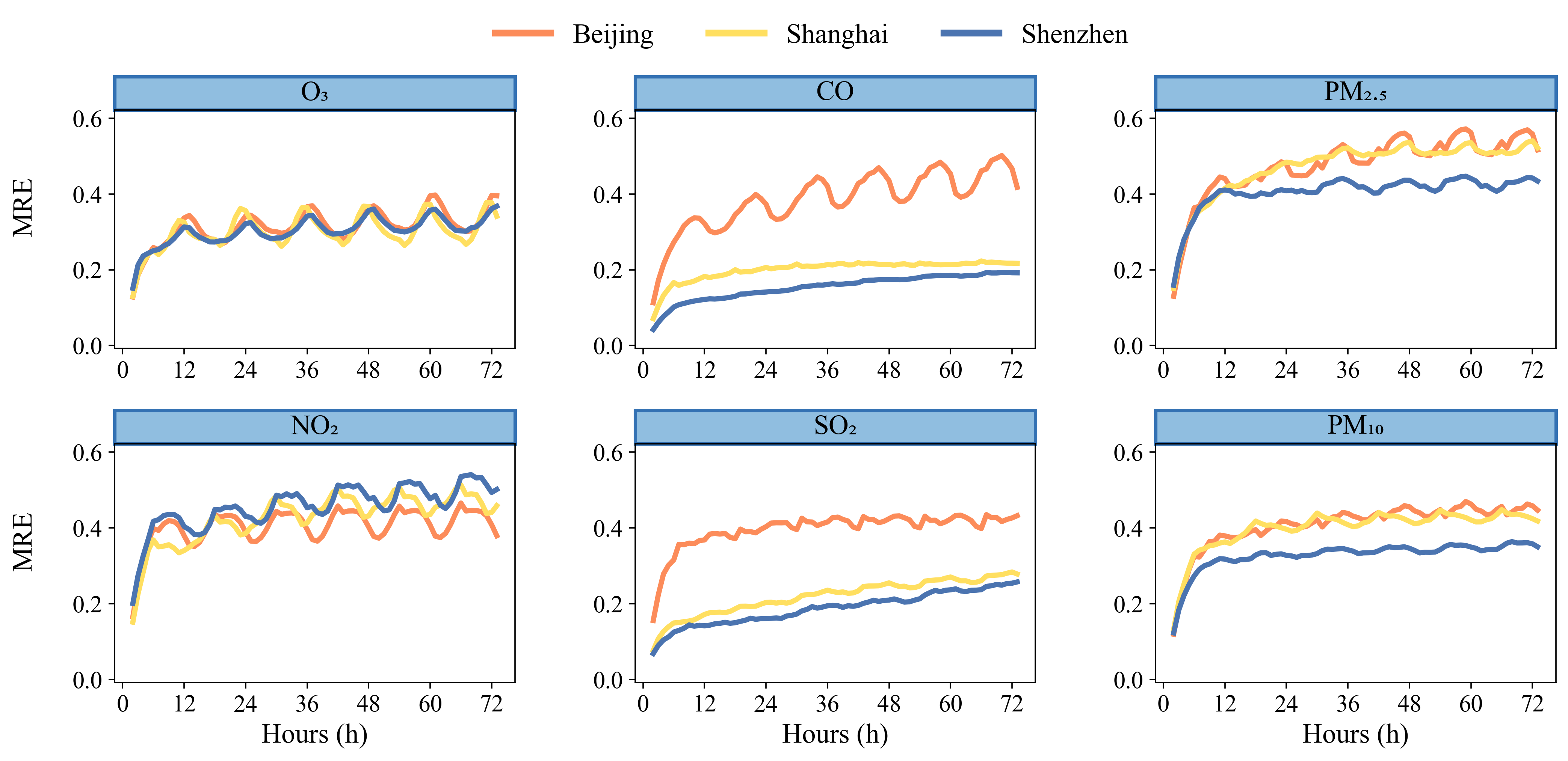}
\caption{MRE evaluation metric curves produced by the FuXi-Air model for six pollutants over a 72-hour forecasting horizon in Shanghai, Beijing, and Shenzhen.}
\label{fig:3.2mre}
\end{figure}

We applied the model to three major cities in China—Beijing, Shanghai, and Shenzhen—and evaluated its predictive performance for six air pollutants over a 72-hour forecasting horizon. The results are presented in Fig. ~\ref{fig:3.2mre} and Table ~\ref{tab:1}. Overall, the model demonstrated high accuracy across all pollutants, with MREs consistently less than 53\%. To facilitate the analysis of simulation differences among the six pollutants, the averages across the three cities were calculated. Among the six pollutants, O$_{3}$ exhibited the best predictive performance, with the associated MREs ranging from 31.27\% to 34.12\% during the 49–72-hour forecasting period. The R reached 0.85, whereas the RMSE and MAE were 26.82 $\mu\text{g/m}^3$ and 20.19 $\mu\text{g/m}^3$, respectively. NO$_{2}$ and PM$_{2.5}$  showed slightly lower but still reliable predictive performance, followed by SO$_{2}$ and CO,for which the model accuracy varied considerably across regions. The PM$_{10}$ predictions were comparatively less accurate, with their MREs ranging from 35\% to 45\%, a maximum RMSE of 87.32 $\mu\text{g/m}^3$, and an MAE of 34.67 $\mu\text{g/m}^3$. In terms of temporal evolution, the prediction errors induced by the model increased more rapidly within the first 24 hours, as indicated by more pronounced fluctuations in the corresponding error curves. However, beyond 25 hours, the errors tended to stabilize, suggesting that the model effectively captured short-term concentration variations and maintained robustness over longer forecasting periods.

From the perspective of pollutant types, the model demonstrated greater stability in predicting secondary pollutants such as O$_{3}$, with relatively small MRE fluctuations. In contrast, although the overall errors observed for primary pollutants such as SO$_{2}$ and CO remained low, their predictions showed greater variability. For example, the MRE of O$_{3}$ decreased to 24.90\% at the 6th forecasting step and remained stable at approximately 32\% over the subsequent 66 steps, with a fluctuation range of less than 5\%. Interestingly, NO$_{2}$, —an important precursor of O$_{3}$ —also exhibited strong stability: during the first 24 steps, its MRE was 38.66\%, with an R of 0.71, and the maximum variation in the MRE across the subsequent 48 steps was only 4.68\%. These results indicate that the model successfully captured the coupling relationship between O$_{3}$ and NO$_{2}$,. As a typical secondary pollutant formed through photochemical reactions, O$_{3}$ concentrations are significantly influenced by meteorological conditions such as temperature, as well as precursor substances such as nitrogen oxides\cite{coates2016influence,seinfeld2016atmospheric}. The forecasts of NO$_{2}$, and O$_{3}$ displayed a certain degree of synchronicity, which can be attributed to the photochemical equilibrium formed between the titration effect of NO$_{2}$, and the generation of O$_{3}$. Since the model fully learned the influences of meteorological conditions and precursor substances on O$_{3}$ concentrations during the training phase, it was able to accurately predict O$_{3}$ concentration variations during the forecasting stage. This further validates the accuracy and stability of the model in terms of simulating secondary pollutants. In contrast, SO$_{2}$ and CO presented greater variability in their MRE and R values (MRE: 17.78–43.67\%, R: 0.28–0.81), indicating a certain degree of uncertainty in the responses of the model to these pollutants. Additionally, PM$_{2.5}$ maintained a consistently high correlation throughout the 72-hour forecasting period, with its R values remaining above 0.60 and its RMSEs always below 27.23 $\mu\text{g/m}^3$, demonstrating strong fitting performance. However, the prediction errors induced for PM$_{10}$ were relatively large (MRE$_{avg}$ = 46.97\%, RMSE$_{avg}$ = 87.32 $\mu\text{g/m}^3$, R$_{avg}$ = 0.39), primarily due to its concentrations being more susceptible to regional transport and episodic events such as dust storms. These factors introduced uncertainties that reduced the responsiveness of the model to abrupt concentration surges, suggesting that its ability to capture such transient spikes requires further improvement.

\begin{table}[H]
    \centering
    \scriptsize
    \renewcommand{\arraystretch}{1.3} 
    \setlength{\tabcolsep}{1.2pt}  
        \caption{MREs,MAEs, RMSEs, and R values are reported for the six main pollutants in different cities, where the unit of RMSEs and MAEs for CO is mg/m³, while for the other pollutants it is $\mu\text{g/m}^3$; MREs and R are unitless.}
    \label{tab:1}
\begin{tabular}{cccccccccccccc}
\hline
\multicolumn{2}{c}{City}        & \multicolumn{4}{c}{Beijing} & \multicolumn{4}{c}{Shanghai} & \multicolumn{4}{c}{Shenzhen} \\ \cline{3-14} 
\multicolumn{2}{c}{Indicator}   & MREs  & MAEs   & RMSEs  & R    & MREs   & MAEs   & RMSEs  & R    & MREs   & MAEs   & RMSEs  & R    \\ \hline
\multirow{3}{*}{O$_{3}$}    & 1-24h  & 0.28 & 17.76 & 24.07 & 0.88 & 0.28  & 18.82 & 25.36 & 0.81 & 0.27  & 16.23 & 22.53 & 0.81 \\
                       & 25-48h & 0.32 & 19.76 & 26.20  & 0.86 & 0.31  & 20.88 & 27.42 & 0.78 & 0.31  & 18.31 & 24.44 & 0.77 \\
                       & 49-72h & 0.34 & 20.55 & 27.31 & 0.85 & 0.31  & 21.00    & 27.77 & 0.77 & 0.32  & 19.03 & 25.38 & 0.76 \\
\multirow{3}{*}{NO$_{2}$}   & 1-24h  & 0.38 & 9.75  & 13.77 & 0.77 & 0.36  & 10.33 & 14.36 & 0.73 & 0.40   & 7.45  & 11.27 & 0.63 \\
                       & 25-48h & 0.41 & 10.76 & 14.84 & 0.73 & 0.45  & 12.72 & 16.89 & 0.64 & 0.47  & 8.52  & 12.30  & 0.54 \\
                       & 49-72h & 0.42 & 10.97 & 15.16 & 0.72 & 0.47  & 13.11 & 17.3  & 0.62 & 0.50   & 8.99  & 12.90  & 0.53 \\
\multirow{3}{*}{SO$_{2}$}   & 1-24h  & 0.35 & 1.01  & 1.50   & 0.45 & 0.17  & 1.15  & 1.78  & 0.69 & 0.14  & 0.65  & 1.00     & 0.81 \\
                       & 25-48h & 0.42 & 1.18  & 1.64  & 0.29 & 0.23  & 1.55  & 2.19  & 0.55 & 0.19  & 0.88  & 1.23  & 0.72 \\
                       & 49-72h & 0.42 & 1.19  & 1.64  & 0.28 & 0.26  & 1.75  & 2.40   & 0.47 & 0.23  & 1.06  & 1.42  & 0.62 \\
\multirow{3}{*}{CO}    & 1-24h  & 0.31 & 0.15  & 0.22  & 0.75 & 0.17  & 0.12  & 0.17  & 0.72 & 0.12  & 0.07  & 0.11  & 0.75 \\
                       & 25-48h & 0.41 & 0.19  & 0.26  & 0.69 & 0.21  & 0.14  & 0.19  & 0.6  & 0.16  & 0.10   & 0.14  & 0.54 \\
                       & 49-72h & 0.44 & 0.21  & 0.28  & 0.68 & 0.22  & 0.14  & 0.20   & 0.57 & 0.19  & 0.12  & 0.16  & 0.39 \\
\multirow{3}{*}{PM$_{2.5}$} & 1-24h  & 0.40  & 13.4  & 23.39 & 0.77 & 0.39  & 10.52 & 15.14 & 0.73 & 0.37  & 6.22  & 8.69  & 0.67 \\
                       & 25-48h & 0.50  & 16.93 & 26.29 & 0.71 & 0.51  & 13.54 & 18.90  & 0.65 & 0.42  & 7.05  & 9.35  & 0.61 \\
                       & 49-72h & 0.53 & 18.13 & 27.23 & 0.70  & 0.52  & 13.69 & 19.10  & 0.63 & 0.43  & 7.21  & 9.68  & 0.60  \\
\multirow{3}{*}{PM$_{10}$}  & 1-24h  & 0.35 & 27.10  & 76.22 & 0.57 & 0.35  & 17.32 & 36.07 & 0.63 & 0.30   & 10.15 & 14.73 & 0.72 \\
                       & 25-48h & 0.43 & 33.25 & 86.04 & 0.44 & 0.42  & 20.81 & 40.10  & 0.45 & 0.34  & 11.55 & 16.22 & 0.65 \\
                       & 49-72h & 0.45 & 34.67 & 87.32 & 0.42 & 0.43  & 21.23 & 40.74 & 0.39 & 0.35  & 12.00    & 17.08 & 0.63 \\ \hline
\end{tabular}

\end{table}

\subsubsection{Model Performance Comparison Across Different Cities}

To systematically evaluate the generalizability of the FuXi-Air model across cities with different pollution profiles, a consistent data processing pipeline, the same model architecture, and the same training methodology were applied in this study to compare the predictive performance achieved for six pollutants over 1–72-hour forecasting windows in Beijing, Shanghai, and Shenzhen. As shown in Table ~\ref{tab:1},the model achieved satisfactory prediction performance in all three cities. However, due to regional pollution mechanism differences, significant variations were observed across the city–pollutant combinations. 

These differences reflect the distinct emission structures, meteorological patterns, and topographic influences at the urban level, further underscoring the importance of treating the city as a fundamental unit in both air quality forecasting and management. In terms of overall performance, Beijing yielded superior performance to that Shanghai and Shenzhen for most pollutants, with the exception of SO$_{2}$ and PM$_{10}$, for which the predictive accuracy of the model was lower. Specifically, the R values achieved for pollutants in Beijing were generally higher than those in the other two cities, indicating superior forecasting performance. From a pollutant perspective, O$_{3}$ demonstrated the highest degree of generalization across cities, with intercity MRE differences of less than 3\%. Both NO$_{2}$, and O$_{3}$ had their lowest errors in Beijing, where the 72-hour MRE for O$_{3}$ was only 37.53\%, and the R values remained above 0.81. Shanghai followed, while Shenzhen had the greatest number of errors—the MREs for NO$_{2}$, and O$_{3}$ reached 54.05\% and 36.78\%, respectively. Nevertheless, all three cities have maintained R values above 0.75 across the full 72-hour forecast. Regarding primary pollutants such as SO$_{2}$ and CO, Beijing presented the highest MRE$_{avg}$ for SO$_{2}$, largely due to its extremely low observed concentration baseline (Beijing: 2.88 $\mu\text{g/m}^3$; Shanghai: 7.10 $\mu\text{g/m}^3$; Shenzhen: 5.10 $\mu\text{g/m}^3$), which led to inflated relative error values under the MRE calculation formula. However, the CO forecasts in Beijing outperformed those in the other two cities, with the associated Ravg values reaching 0.68 during the 49–72-hour forecast window..

 For particulate matter, the error induced by the model exhibited a “higher in the north, lower in the south” trend. The prediction errors observed for PM$_{2.5}$ (RMSE$_{avg}$ = 27.23 $\mu\text{g/m}^3$) and PM$_{10}$ (RMSE$_{avg}$ = 87.32 $\mu\text{g/m}^3$) were greater in Beijing, which may be attributed to the underestimation effect of the model during heavy pollution episodes and the influence of long-range pollutant transport\cite{kong2016empirical,li2016observation}. In contrast, the relatively stable emission conditions of Shenzhen contributed to a significantly lower RMSE$_{avg}$ for PM$_{2.5}$ (9.68 $\mu\text{g/m}^3$), representing a 64.5\% reduction compared to that obtained for Beijing. This finding highlights the model’s robustness under low-variability conditions and its capacity to capture urban-scale pollution dynamics across diverse environments.
 
 Overall, the FuXi-Air model showed good accuracy in the prediction of various pollutants, and achieved stable and high-precision forecasting results in cities with diverse meteorological conditions and emission characteristics. This shows that the model constructed in this study has excellent urban scale generalization capabilities and can effectively cope with modeling challenges under different urban pollution evolution mechanisms. Comparative prediction plots produced for the 24-h, 48-h, and 72-h forecasting horizons applied in the three cities are provided in Supplementary Figs. \ref{Supplementary Fig:beijing_24h}–\ref{Supplementary Fig:shenzhen_72h}.

\subsubsection{Comparisons Between FuXi-Air and Operational Numerical Air Quality Models}  

\begin{figure}[H]
\centering
\includegraphics[width=\linewidth]{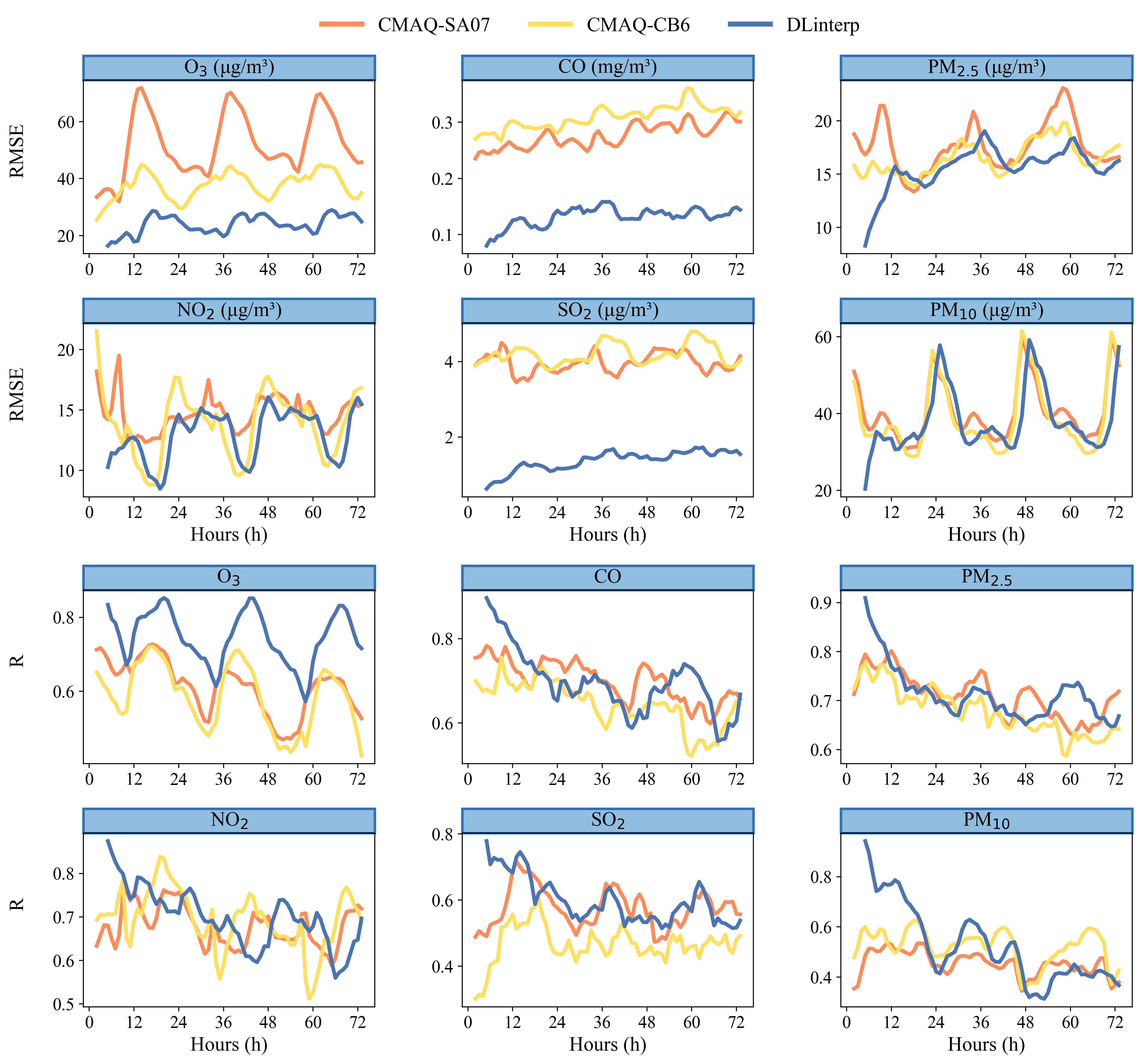}
\caption{Comparison among the 72-hour forecasts of the FuXi-Air model and two numerical models in Shanghai. The SA07 configuration (CMAQ–SA07) adopted a 3-domain nested structure (36–12 km resolution). The CB06 configuration (CMAQ–CB06) applied a finer 4-domain nesting (27–9–3 km resolution). Emission inventories included EDGAR v6.1 (2018) for the global domain, MEIC v1.3 (2019) for China, and 2019YRD100 for the Yangtze River Delta. Biogenic and dust emissions were treated using MEGAN and WBDUST, respectively. Region-specific emissions over Shanghai were provided by the Shanghai Academy of Environmental Sciences.}
\label{fig:3.2.3_DL-CMAQ}
\end{figure}

To evaluate the predictive capability of the proposed FuXi-Air model against traditional numerical approaches, we conducted a comprehensive comparative analysis involving numerical models driven by different atmospheric chemical mechanisms. Specifically, simulations were performed using the Weather Research and Forecasting (WRF, version 4.3.1) model coupled with the Community Multiscale Air Quality (CMAQ, version 5.3.3) model, incorporating two distinct chemical mechanisms: SAPRC07TC-AE6-AQ (SA07) and CB6R3-AE6-AQ (CB06). Forecasting results over a nine-month period from February to December 2023 were evaluated. In terms of computational efficiency, the FuXi-Air model completes a 72-hour air quality forecast for megacities within 25-30 seconds, significantly outperforming traditional numerical models in speed. In contrast, the WRF-CMAQ model typically requires 2 to 3 hours to complete a forecast of the same duration. This substantial time advantage makes the FuXi-Air model much more efficient and practical for operational forecasting. Fig. ~\ref{fig:3.2.3_DL-CMAQ} presents a performance comparison between the FuXi-Air model and numerical models for 72-hour air quality forecasts in the Shanghai region. Due to differences in forecast initialization times between the FuXi-Air and traditional numerical models, the analysis was limited to steps 4-72 for consistency. The results demonstrate that the FuXi-Air has significant advantages in terms of forecasting O$_{3}$, SO$_{2}$, and CO. Compared with those of SA07, the 72-hour RMSEs of the proposed model were reduced by 53.95\%, 65.43\%, and 52.22\%, respectively. Compared with those of CB6, the reductions were 36.21\%, 67.20\%, and 57.28\%, respectively. In terms of R, the FuXi-Air model achieved higher R values than the numerical models did in 66, 40, and 28 of the 69 steps (steps 4-72) for O$_3$, SO$_2$, and CO, respectively, indicating substantial error reductions (Supplementary Fig. ~\ref{Supplementary Fig:3.2.3FuXi-Air_vs._WRF-CMAQ}).For PM$_{2.5}$, NO$_2$, and PM$_{10}$, the 72-hour RMSEs of the FuXi-Air model were 15.62~$\mu$g/m$^3$, 12.92~$\mu$g/m$^3$, and 38.00~$\mu$g/m$^3$, representing decreases of 11.29\%, 12.09\%, and 6.41\%, respectively, compared with those of SA07 and decreases of 5.30\%, 6.29\%, and 2.41\% compared with those of CB6. Overall, the  FuXi-Air model outperformed the traditional numerical models in terms of the prediction accuracy achieved for all six major pollutants across both short-term (steps 4-24) and medium- to long-term (steps 24-72) forecasts, demonstrating strong generalizability and robustness.

\subsection{Interpretability Analysis of the FuXi-Air Forecasting Results} 

\subsubsection{Influences of Meteorological Conditions and Emission Inventories on the Model Predictions}  

\begin{figure}[H]
\centering
\includegraphics[width=\linewidth]{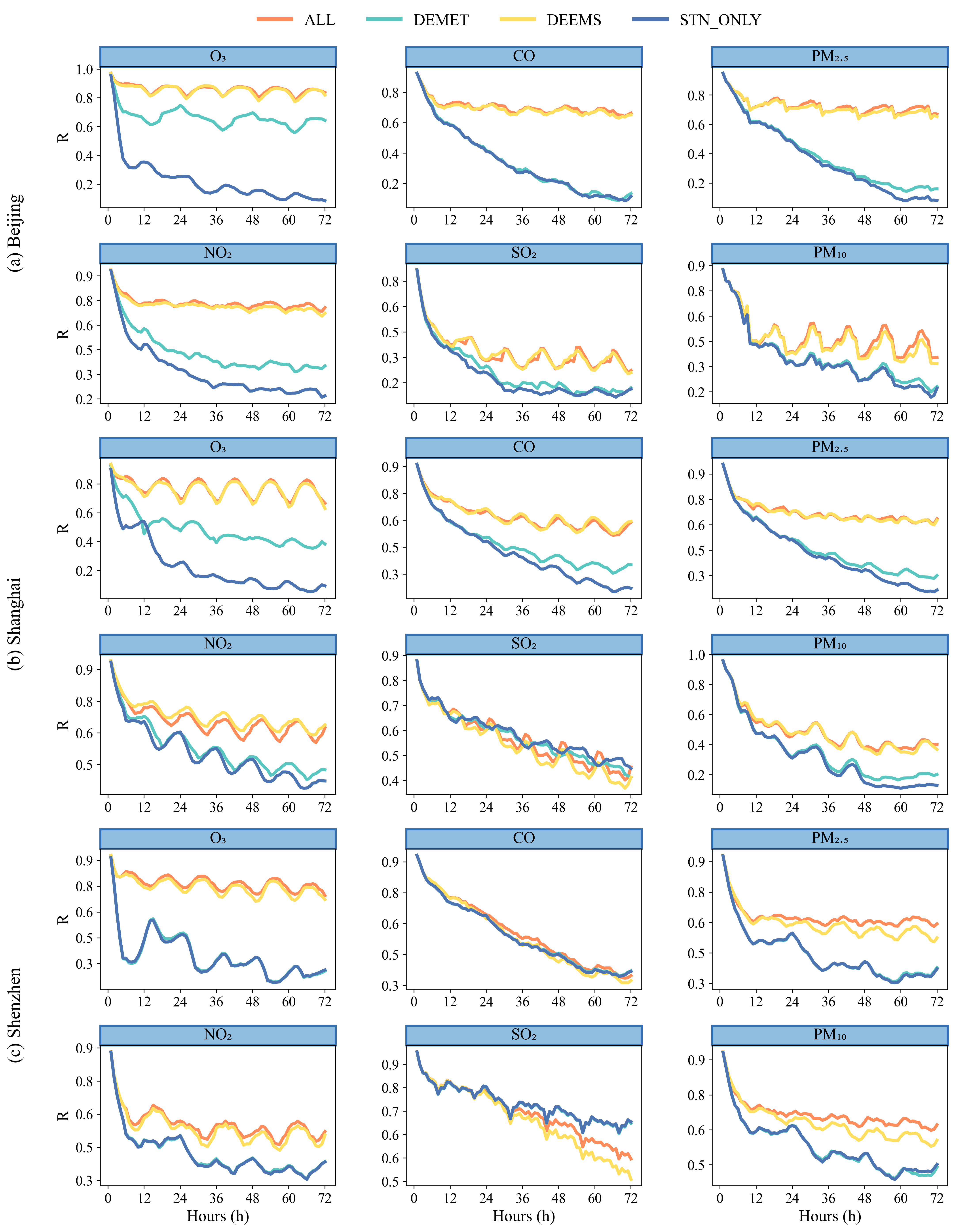}
\caption{R curves derived from ablation experiments involving Beijing, Shanghai, and Shenzhen.}
\label{fig:3.3R}
\end{figure}

Meteorological conditions and emission sources jointly govern the spatiotemporal dynamics of air pollutant concentrations. In this study, four ablation experiments (Table ~\ref{tab:Supplementary Table 2} in the Dataset Introduction section) were conducted to systematically evaluate the individual and combined contributions of meteorological data, emission inventories, and station observations to the prediction accuracies achieved for six pollutants (PM$_{2.5}$, PM$_{10}$, O$_{3}$, NO$_{2}$, CO, and SO$_{2}$) across Beijing, Shanghai, and Shenzhen.The experimental results (Figs. ~\ref{fig:3.3R} and ~\ref{fig:3.3Boxplots}, Supplementary Fig. ~\ref{Supplementary Fig:rmse}, Supplementary Table    ~\ref{tab:Supplementary Table 3}) show that the ALL group—which integrated multisource inputs (meteorological, emission, and station data)—achieved the best predictive performance. For key pollutants such as O$_{3}$ and PM$_{2.5}$, the average correlation coefficient (R$_{avg}$) across the three cities in the 1–72-hour forecasting windows ranged from 0.63 to 0.86, with the RMSEs significantly reduced in 84\% of the experiments. In contrast, the \textnormal{STN\_ONLY} group (Fig. ~\ref{fig:3.3R}, \textnormal{STN\_ONLY} Group), which relied solely on station observation data, exhibited pronounced limitations, with $R_{\text{avg}}$ values below 0.5 and over 95\% of the RMSE values reaching their highest levels. These findings confirm that multisource data fusion is critical for improving the predictive accuracy of the model.

A controlled variable analysis revealed that meteorological data contributed significantly more to improving the performance of the model than emission inventories did. This advantage was particularly evident for pollutants that are sensitive to photochemical reactions and regional transport, such as O$_{3}$, NO$_{2}$, PM$_{2.5}$, and PM$_{10}$. For example, within the 1–72-hour forecasting windows, incorporating meteorological data (Fig. ~\ref{fig:3.3R}, DEEMS Group;  Figs. ~\ref{fig:3.3Boxplots} (a) (b) panels) increased the R$_{avg}$ value of the O$_{3}$ prediction results obtained in Beijing from 0.22 to 0.86 ($\Delta R_{\mathrm{avg}}$ = 0.64), while reducing the $RMSE_{\text{avg}}$ by 23.38 $\mu\text{g/m}^3$. In comparison, the inclusion of the emission inventory alone  (Fig. ~\ref{fig:3.3R}, DEMET Group;  Figs. ~\ref{fig:3.3Boxplots} (e) (f) panels)  resulted in a smaller improvements ($\Delta R_{\mathrm{avg}}$ = 0.44, $\Delta RMSE_{\mathrm{avg}}$ = 11.04 $\mu\text{g/m}^3$). These findings highlight the critical role of meteorological variables in capturing pollution trends and peak responses.

The contributions of meteorological conditions and emission inventories to pollutant concentration forecasting exhibit multidimensional heterogeneity, varying with the input data configurations and pollutant types. When the model was fed with limited inputs, the inclusion of meteorological data (Fig. ~\ref{fig:3.3R}, DEEMS vs. \textnormal{STN\_ONLY} groups) leads to a maximum R improvement of 0.64, effectively compensating for deficiencies in modeling regional transport and photochemical processes. In contrast, under multisource data fusion scenarios(Fig. ~\ref{fig:3.3R}, ALL vs. DEMET groups; Figs. ~\ref{fig:3.3Boxplots} (c) (d) panels), the contribution of meteorological data to concentration forecasting exhibited a clear attenuation trend ($\Delta R_{\mathrm{avg}}$\_max = 0.45), likely due to the synergistic effects that were already captured by the combination of emission inventories and meteorological factors. A similar pattern was observed for the predictive gain derived from emission inventories. In the absence of meteorological data (Fig. ~\ref{fig:3.3R}, DEMET vs. \textnormal{STN\_ONLY} groups; Figs. ~\ref{fig:3.3Boxplots} (e) (f) panels), incorporating emission inventories could improve R by up to 0.44. However, when meteorological drivers were already present (Fig. ~\ref{fig:3.3R}, ALL vs.DEEMS groups; Figs. ~\ref{fig:3.3Boxplots} (g) (h) panels), the improvement significantly diminished ($\Delta R_{\mathrm{avg}}$\_max = 0.05). These results suggest that pollutant concentrations are jointly influenced by meteorology and emissions, with meteorological conditions playing a dominant role in shaping concentration trends. This influence operates through physical processes such as transport and dispersion, as well as by enabling chemical processes such as photochemical reactions via temperature and solar radiation—partially masking the local effects of emission sources.

A pollutant-specific analysis revealed significant differences among the responses of distinct species to driving factors (Fig. ~\ref{fig:3.3R},Figs. ~\ref{fig:3.3Boxplots}). Meteorological variables exert a decisive influence on the prediction of secondary pollutant concentrations (O$_{3}$ and NO$_{2}$). After incorporating meteorological factors, the improvements in the $R_{\mathrm{avg}}$ and the reductions in the $rRMSE_{\mathrm{avg}}$ for secondary pollutants are significantly greater than those observed for primary pollutants such as CO and SO$_{2}$. This confirms that the secondary pollutant formation process critically depends on the nonlinear chemical mechanisms modulated by meteorological parameters (e.g., temperature and humidity), whereas the primary pollutant concentrations are predominantly governed by the emission source intensity. For PM$_{2.5}$ and PM$_{10}$ as mixed-type pollutants, their predictive performance was codetermined by both meteorological conditions and emission sources, reflecting the hybrid nature of their formation pathways.

Overall, meteorological data are essential for constructing high-precision pollutant concentration forecasting models. The inclusion of meteorological conditions is critical to ensuring the ability of the model to capture both concentration trends and peak responses. In contrast, the emission inventory enhances the capacity of the model to resolve local emission characteristics through its nonlinear interactions with meteorological drivers. The coupling of these two components significantly improves the overall ability of the model to simulate complex atmospheric pollution processes.

\subsubsection{Effectiveness of FuXi Meteorological Forecasts as Model Drivers}  

For the first time, a pollutant forecasting framework coupled with a high-resolution FuXi-2.0 meteorological prediction system was established in this study. Through a pollutant–emission inventory–meteorology coupling module, the framework enables the offline integration of site-level observations, emission inventory data, and hourly forecasts derived from the AI-driven FuXi model. The system supports forecasting up to 72 hours ahead, with a temporal resolution of 1 hour. Comparative experiments (Table ~\ref{tab:2}) show that within the 49–72-hour forecasting window, the FuXi-driven model yielded  $RMSE_{\mathrm{avg}}$ increases of 0.51 $\mu\text{g/m}^3$, 0.01 $\text{mg/m}^3$, and 0.29 $\mu\text{g/m}^3$ for PM$_{10}$, CO, and SO$_{2}$, respectively, relative to the ERA5 reanalysis data. For O$_{3}$, NO$_{2}$, and PM$_{2.5}$, the $RMSE_{\mathrm{avg}}$ differences were all within 3 $\mu\text{g/m}^3$ ($\Delta R_{\mathrm{avg}}$ < 0.06, $\Delta MRE_{\mathrm{avg}}$ < 0.09). These results indicate that the forecasting accuracy of the new-generation AI-based FuXi-2.0 model approaches the equivalent level of traditional reanalysis products (ERA5), offering a novel and reliable meteorological driving solution for operational air quality forecasting.

\begin{figure}[H]
\centering

\includegraphics[width=\linewidth]{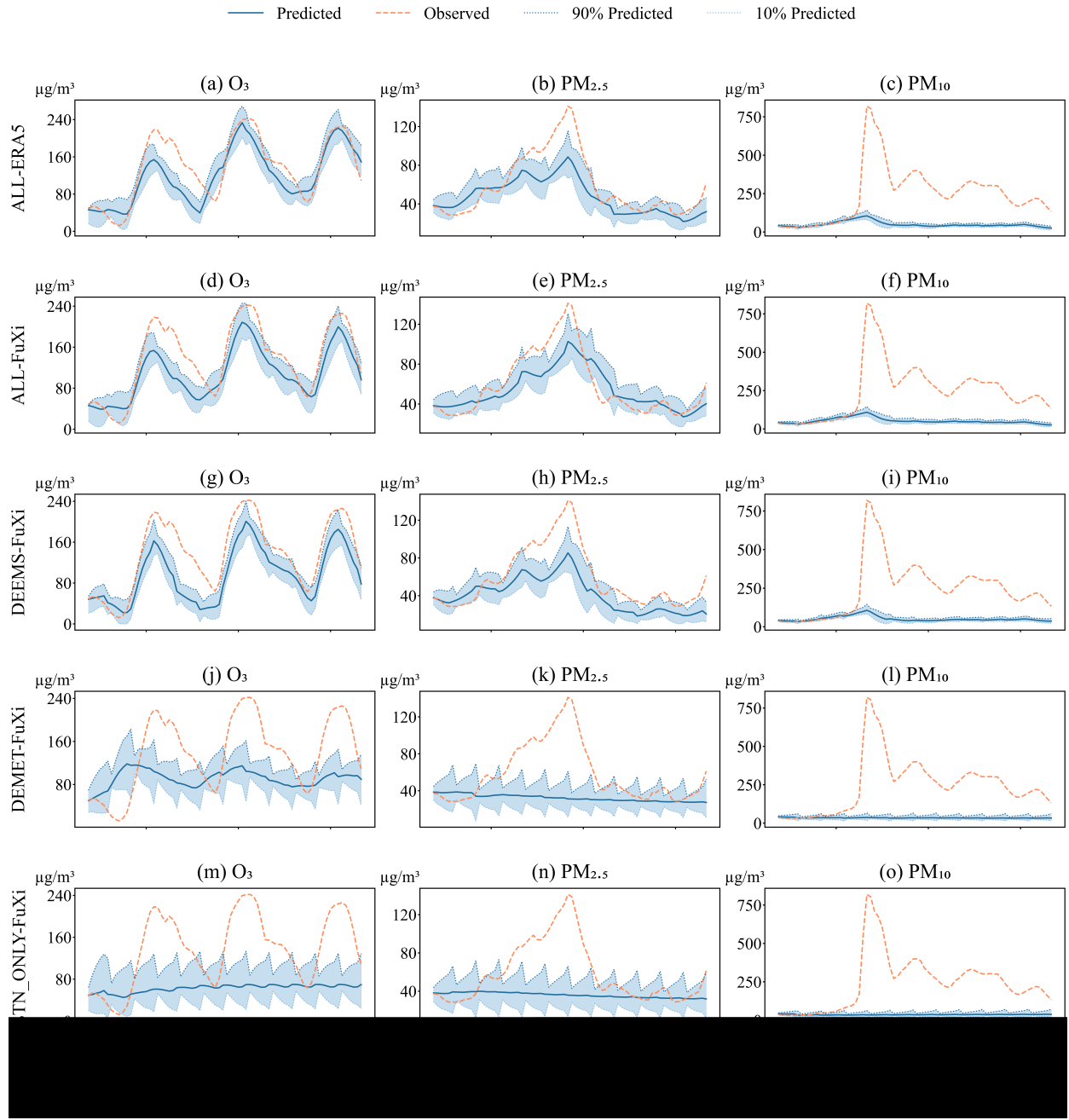}

\caption{Time series forecasts produced for pollution episodes in Shanghai by different model configurations (timestamps in China Standard Time, CST)}
\label{fig:3.3.2Forecasts}
\end{figure}

Further ablation analyses of multimodal data and the embedded physicochemical mechanisms employed during forecasting (Fig. ~\ref{fig:3.3.2Forecasts}) revealed that for PM$_{2.5}$ pollution events, the pollutant accumulation process occurring under stagnant meteorological conditions was highly dependent on atmospheric variables. Compared with the outcomes in Fig. ~\ref{fig:3.3.2Forecasts} (e), removing the meteorological data (Fig. ~\ref{fig:3.3.2Forecasts} (k)) resulted in the model failing to capture the periodic fluctuations and peak occurrences exhibited by the PM$_{2.5}$ concentrations, leading to a smoothed and flattened forecasting sequence. Excluding the emission inventory (Fig. ~\ref{fig:3.3.2Forecasts} (h)) caused a decrease in the simulation accuracy and a reduction in the predicted peak values. Similarly, on O$_{3}$ pollution days, the absence of meteorological data (Fig. ~\ref{fig:3.3.2Forecasts} (j)) disrupted the photochemical chain reaction mechanism, causing both the diurnal peaks and daily variation patterns to deviate substantially from the observations. Removing the emission inventory (Fig. ~\ref{fig:3.3.2Forecasts} (g)) also led to a noticeable decline in the simulation accuracy of the model.

In addition, the regional transport characteristics of PM$_{10}$ revealed the implicit learning capabilities of the model. Even without the explicit incorporation of external pollution sources, the interactions between the meteorological conditions (e.g., wind direction and humidity) and local site concentrations (Figs. ~\ref{fig:3.3.2Forecasts} (f) (i) panels) could still weakly reflect cross-regional transport patterns. However, when the meteorological data were removed (Figs. ~\ref{fig:3.3.2Forecasts} (l) (o) panels), such responses were entirely eliminated, indicating that meteorological inputs implicitly encode the key information related to pollutant transport pathways.

In summary, FuXi meteorological forecasts have clear advantages. The prediction framework based on the FuXi AI meteorological model already exhibits a forecasting performance that is comparable to that of traditional reanalysis-driven systems. The reliability of pollutant concentration forecasts depends on the accurate representation of dynamic processes by meteorological data, the spatiotemporal representativeness of the emission inventory, and the real-time correction capability provided by station observations.As the fundamental unit of air quality management, cities are subject to locally specific redistribution and response patterns even when pollutants are transported from regional sources. The current capacity of city-scale modeling can, to a certain extent, capture these transport processes, indicating the model’s capability to resolve urban-level pollution evolution dynamics.

\begin{table}[H]
    \centering
    \scriptsize
    \renewcommand{\arraystretch}{1.3} 
    \setlength{\tabcolsep}{1.2pt}  
    \caption{Comparison among the evaluation metrics of models trained with ERA5 and FuXi meteorological data, where the unit of RMSEs for CO is mg/m³, while for the other pollutants it is $\mu\text{g/m}^3$; MREs and R are unitless.}
    \label{tab:2}
\begin{tabular}{cccccccc}
\hline
\multicolumn{2}{c}{Indicator}                 & \multicolumn{2}{c}{R} & \multicolumn{2}{c}{RMSEs} & \multicolumn{2}{c}{MREs} \\ \hline
\multicolumn{2}{c}{Meteorological Data Types} & ERA5      & FuXi      & ERA5        & FuXi       & ERA5       & FuXi       \\ \hline
\multirow{3}{*}{O$_{3}$}           & 1-24h         & 0.83      & 0.81      & 23.36       & 25.36      & 0.26       & 0.28       \\
                              & 25-48h        & 0.80      & 0.78      & 25.24       & 27.42      & 0.29       & 0.31       \\
                              & 49-72h        & 0.80      & 0.77      & 25.45       & 27.78      & 0.29       & 0.31       \\
\multirow{3}{*}{NO$_{2}$}          & 1-24h         & 0.76      & 0.73      & 13.34       & 14.36      & 0.32       & 0.36       \\
                              & 25-48h        & 0.70      & 0.64      & 14.82       & 16.89      & 0.38       & 0.45       \\
                              & 49-72h        & 0.68      & 0.62      & 15.22       & 17.30      & 0.40       & 0.47       \\
\multirow{3}{*}{PM$_{2.5}$}        & 1-24h         & 0.74      & 0.73      & 14.23       & 15.14      & 0.36       & 0.39       \\
                              & 25-48h        & 0.68      & 0.65      & 16.05       & 18.90      & 0.42       & 0.51       \\
                              & 49-72h        & 0.67      & 0.63      & 16.22       & 19.10      & 0.43       & 0.52       \\
\multirow{3}{*}{PM$_{10}$}         & 1-24h         & 0.63      & 0.63      & 36.05       & 36.07      & 0.34       & 0.35       \\
                              & 25-48h        & 0.45      & 0.45      & 40.01       & 40.10      & 0.39       & 0.42       \\
                              & 49-72h        & 0.41      & 0.39      & 40.23       & 40.74      & 0.40       & 0.43       \\
\multirow{3}{*}{CO}           & 1-24h         & 0.71      & 0.72      & 0.16        & 0.17       & 0.17       & 0.17       \\
                              & 25-48h        & 0.61      & 0.60      & 0.19        & 0.19       & 0.20       & 0.21       \\
                              & 49-72h        & 0.57      & 0.57      & 0.19        & 0.20       & 0.21       & 0.22       \\
\multirow{3}{*}{SO$_{2}$}          & 1-24h         & 0.70      & 0.69      & 1.73        & 1.78       & 0.16       & 0.17       \\
                              & 25-48h        & 0.59      & 0.55      & 1.98        & 2.19       & 0.20       & 0.23       \\
                              & 49-72h        & 0.53      & 0.47      & 2.12        & 2.41       & 0.22       & 0.26       \\ \hline
\end{tabular}
\end{table}

\subsubsection{Interpretation of Intercity Air Pollution Mechanism Differences}  

\begin{figure}[H]
\centering
\includegraphics[width=\linewidth]{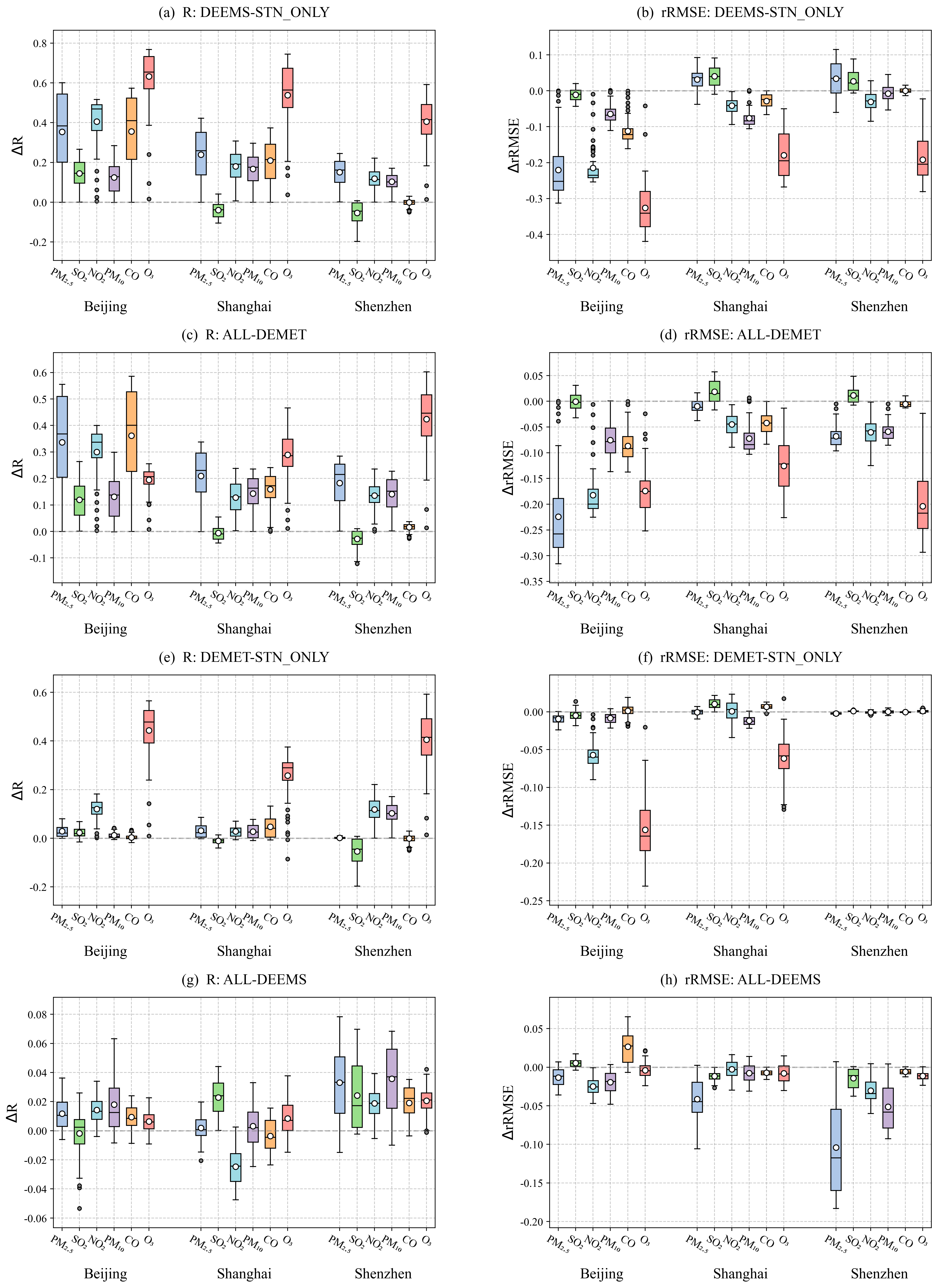}
\caption{Differences among the evaluation metrics produced across different ablation experimental groups.
Panels (a) and (b): DEEMS vs. STN\_ONLY; (c)(d): ALL vs. DEMET; (e)(f): DEMET vs. STN\_ONLY; and (g)(h): ALL vs. DEEMS.}
\label{fig:3.3Boxplots}
\end{figure}

A comparative analysis conducted based on multicity ablation experiments revealed that the simulation performances achieved for O$_{3}$ and its precursor NO$_x$ exhibited distinct regional sensitivities to meteorological forcing(Fig. ~\ref{fig:3.3Boxplots}). An evaluation conducted using increments in the correlation coefficient ($\Delta R_{\mathrm{avg}}$)and reductions in the relative RMSE ($\Delta rRMSE_{\mathrm{avg}}$) revealed that the inclusion of meteorological data improved the $\Delta R_{\mathrm{avg}}$ values obtained for O$_{3}$ and NO$_x$ by 0.12–0.64 and reduced rRMSEs by 0.03–0.33, and the obtained—improvements were significantly greater than those observed for other pollutants.

Regionally, under the single-source data scenario, meteorological inputs led to the greatest improvements in O$_{3}$/NO$_x$ prediction improvements for Beijing ($\Delta R_{\mathrm{avg}}$ = 0.63/0.41), followed by Shanghai ($\Delta R_{\mathrm{avg}}$ = 0.54/0.18), and then Shenzhen ($\Delta R_{\mathrm{avg}}$ = 0.41/0.12)(Figs. ~\ref{fig:3.3Boxplots} (a)(b) panels).However, under the multisource data fusion scenario (Figs. ~\ref{fig:3.3Boxplots} (c)(d) panels), the benefit of the meteorological input was most pronounced in Shenzhen, where the $\Delta R_{\mathrm{avg}}$ for O$_{3}$ reached 0.42 and the $\Delta rRMSE_{\mathrm{avg}}$ decreases by 0.2. This indicates that meteorological drivers yielded a stronger contribution when interacting with emissions, suggesting that in emission-driven cities, the formation of secondary pollutants is governed by pronounced meteorology–emissions coupling. Achieving improved predictive accuracy in such cases depends on the synergistic optimization of multisource data.

An analysis of the driving mechanisms of mixed-type pollutants (PM$_{2.5}$ and PM$_{10}$) revealed that their concentrations were jointly controlled by meteorological transport and local emissions, with strong regional variations. In Beijing, the PM forecasting performance was governed primarily by meteorological factors due to cross-border dust transport driven by the Mongolian High and wintertime coal combustion for heating\cite{yin2025regional} . This was evidenced by the substantial performance gains attained when meteorological data were included (Figs. ~\ref{fig:3.3Boxplots} (c)(d) panels: $\Delta R_{\mathrm{avg}}$ = 0.13/0.34 for PM$_{10}$/PM$_{2.5}$; $\Delta rRMSE_{\mathrm{avg}}$ = 0.08/0.22), whereas the standalone contribution of the CAMS emission inventory was limited (Figs. ~\ref{fig:3.3Boxplots} (e)(f) panels: $\Delta R_{\mathrm{avg}}$ $ \leq 0.03$  ).

In Shenzhen, a representative industrial city in southern China, the subtropical climate—with high temperature and high humidity—intensified the accumulation of locally emitted pollutants. Consequently, local emissions exerted a stronger influence on PM pollution, as reflected in a PM$_{2.5}$ $\Delta rRMSE_{\mathrm{avg}}$ reduction of 0.10 when emissions were included(Fig. ~\ref{fig:3.3Boxplots} (g) vs. (h) panels). Shanghai lies between these two extremes. Located in the Yangtze River estuary, its PM pollution dynamics are shaped by land–sea breeze circulation and reflect a hybrid control pattern that depends on both meteorological and emission-related inputs.

An analysis of the primary pollutants (CO and SO$_{2}$)further highlighted the impacts of regional pollution mechanisms on model sensitivity. Although their concentrations were largely emission-driven, the meteorological factors still provided notable forecasting improvements in Beijing due to the blocking effect of the Yanshan Mountains and the enhanced atmospheric stability observed during the heating season (Fig. ~\ref{fig:3.3Boxplots} (c) vs. (d) panels: $\Delta R_{\mathrm{avg}}$ = 0.12–0.30). In contrast, the meteorological contribution in Shenzhen was minimal, while the inclusion of emission inventories led to substantial improvements in the R values, underscoring the dominant role of local sources. These differences suggest that the predictive accuracy achieved for primary pollutants depends on the spatial heterogeneity of the emission source distributions, the background meteorological conditions, and the atmospheric dispersion capacity of each city.

In summary, the city-scale experimental comparisons revealed distinct pollution formation mechanism differences. Beijing, as a meteorology-sensitive city, was dominated by pollutant transport under the Siberian High and photochemical processes, with the model performance highly reliant on high-resolution meteorological inputs. Shanghai, which is located in a land–sea interaction zone, experienced coupled accumulation–dispersion dynamics that were influenced by both meteorology and emissions. Shenzhen, as an emission-sensitive city, experienced rapid photochemical reactions under tropical climatic conditions, resulting in shorter pollutant lifetimes and stronger model dependence on emission inventories. These city-specific differences were clearly reflected in the ablation experiments, reinforcing the necessity and applicability of multisource data fusion in city-scale air quality modeling scenarios.

\section{Conclusion and Outlook}  

A meteorology–emission–pollutant coupled air quality forecasting model (FuXi-Air) , which integrates FuXi-2.0 high-resolution meteorological forecasts into a attention-based framework to achieve real-time, highly precise, hourly predictions for multiple pollutants at the urban scale, was proposed in this study. The model effectively addresses the limitations of the traditional numerical simulation approaches, such as their low efficiency, poor generalizability, and high computational costs.

The validation results obtained from Beijing, Shanghai, and Shenzhen indicate that, at the maximum forecasting horizon of 72 hours, the mean relative errors (MREs) induced for major pollutants such as PM$_{2.5}$ and O$_{3}$ consistently remained below 53\%, with the correlation coefficient (R) reaching 0.85, demonstrating the strong ability of the model to accurately predict the concentrations of secondary atmospheric pollutants. Moreover, the model exhibited notable performance differences across different city–pollutant combinations, reflecting city-specific pollution formation mechanisms and urban environmental heterogeneity. This suggests that FuXi-Air is capable not only of capturing regionally distinct pollution dynamics but also of adapting to the pollution characteristics of individual cities, which serve as fundamental units for urban air quality management.

Ablation experiments further confirmed that multisource data fusion yielded optimal predictive performance. The coupling of meteorological inputs and emission inventories significantly enhanced the ability of the model to simulate complex atmospheric pollution processes. Region-specific differences observed in controlled experiments underscore the necessity of integrating heterogeneous data sources into forecasting models. The model’s current performance under city-scale settings has demonstrated a certain capacity to respond to and reflect regional pollutant transport processes as they evolve within urban environments. This implies that even when pollution is driven by large-scale regional transport, its city-level redistribution and manifestation can still be captured to a meaningful degree by the current modeling framework.

Within the 49–72-hour forecasting window, the RMSE variations exhibited by the six target pollutants remained within controllable ranges by FuXi-Air model, providing a reliable meteorological forcing strategy for the operational deployment of AI-based weather models in air quality forecasting tasks. In addition, the unified coupling framework of the model—integrating meteorological data, emission inventories, and pollutant monitoring data—enabled accurate forecasting results to be obtained across urban clusters with diverse pollution mechanisms. This study lays a solid technical foundation for performing data assimilation–based operational air quality forecasting and offers theoretical guidance for developing high-precision AI forecasting systems that combine mechanistic modeling with data-driven methods.

Several key directions for further improvement in near future are identified. In terms of emission inventory optimization, the current inventory is updated on a monthly average basis and suffers from temporal delays. Moreover, the modeled inventory data include only eight precursor species across 17 economic sectors. Future work should aim to incorporate satellite-derived inversion data along with real-time traffic and industrial monitoring data to construct a dynamic emission inventory with a high spatiotemporal resolution and broader sectoral coverage, thereby improving the responsiveness of the model to short-term emission variations. With respect to regional transport modeling, the current single-city-scale forecasting framework is insufficient for capturing large-scale transboundary processes, such as dust transport events (PM$_{10}$). To address this limitation, a multiscale, regionally nested architecture will be developed to resolve cross-boundary transport mechanisms while preserving the specificity of city-level responses. This multiscale architecture will also better capture how transported pollutants interact with local meteorological and emission conditions, improving interpretability and model performance. In terms of the forecasting horizon, the current FuXi meteorological forecasts have already achieved sub-seasonal prediction capabilities, providing a solid foundation for longer-term air pollution forecasting. In the future, the model can be further enhanced by integrating additional multisource data—such as satellite remote sensing data and land-use information—to improve its generalizability and forecasting accuracy, thereby promoting its broader applicability at the national scale. These advancements will further improve the capacity of the model to resolve complex atmospheric pollution processes and provide more reliable technical support for operational air quality forecasting.

\section*{Data Availability Statement}

The CAMS-GLOB-ANT emission inventory data can be retrieved from https://eccad.aeris-data.fr/.
The ERA5 reanalysis data are accessible through the Copernicus Climate Data Store at https://cds.climate.copernicus.eu/.
FuXi-2.0 forecast data are available through the joint data services provided by Fudan University and the Shanghai Academy of AI for Science.
All air pollutant concentration monitoring data were obtained from the China National Environmental Monitoring Center (CNEMC).

\section*{Author contributions statement}

Zhixin Geng: Methodology; Software; Formal analysis; Investigation; Data curation; Writing—original draft preparation; Visualization. Xu Fan: Methodology; Software; Formal analysis; Investigation; Data curation; Writing—original draft preparation; Visualization. Xiqiao Lu: Methodology; Software; Formal analysis; Investigation; Data curation; Writing—original draft preparation; Visualization. Yan Zhang: Conceptualization; Methodology; Resources; Writing—review and editing; Supervision. Guangyuan Yu: Writing—review and editing. Cheng Huang: Resources, Field measurement data, Writing—review and editing. Qian Wang: Resources, Field measurement data. Yuewu Li: Resources, Field measurement data, Investigation; Writing—review and editing. Weichun Ma: Methodology. Qi Yu: Methodology. Libo Wu: Resources; Methodology. Hao Li: Writing—Methodology; Writing—review and editing, Resources, Supervision.

\section*{Acknowledgements}

We sincerely thank the researchers at the Copernicus Atmosphere Monitoring Service (CAMS) and the European Centre for Medium-Range Weather Forecasts (ECMWF) for their valuable contributions in collecting, archiving, disseminating, and maintaining the CAMS-GLOB-ANT emission inventory and ERA5 reanalysis datasets. We are also grateful to Fudan University and the Shanghai Academy of AI for Science for their efforts in collecting, archiving, and maintaining the FuXi-2.0 meteorological forecast data. Our appreciation extends to the China National Environmental Monitoring Center and the Shanghai Environmental Monitoring Center for providing atmospheric pollutant monitoring data. This work is supported by the Project funded by the National Natural Science Foundation of China (42375100), the Natural Science Foundation of Shanghai Committee of Science and Technology, China (22ZR1407700) and 2024 AI for Science Project funded by Fudan University.

\section*{Competing interests}

The authors declare no conflict of interest.

\bibliography{main} 

\begin{thebibliography}{10}
\urlstyle{rm}
\expandafter\ifx\csname url\endcsname\relax
  \def\url#1{\texttt{#1}}\fi
\expandafter\ifx\csname urlprefix\endcsname\relax\def\urlprefix{URL }\fi
\expandafter\ifx\csname doiprefix\endcsname\relax\def\doiprefix{DOI: }\fi
\providecommand{\bibinfo}[2]{#2}
\providecommand{\eprint}[2][]{\url{#2}}

\bibitem{1chen2021possible}
\bibinfo{author}{Chen, S.-L.}, \bibinfo{author}{Chang, S.-W.}, \bibinfo{author}{Chen, Y.-J.} \& \bibinfo{author}{Chen, H.-L.}
\newblock \bibinfo{journal}{\bibinfo{title}{Possible warming effect of fine particulate matter in the atmosphere}}.
\newblock {\emph{\JournalTitle{Communications Earth \& Environment}}} \textbf{\bibinfo{volume}{2}}, \bibinfo{pages}{208} (\bibinfo{year}{2021}).

\bibitem{2xu2025rising}
\bibinfo{author}{Xu, X.}, \bibinfo{author}{Huang, L.}, \bibinfo{author}{Yao, L.}, \bibinfo{author}{Yoshida, Y.} \& \bibinfo{author}{Long, Y.}
\newblock \bibinfo{journal}{\bibinfo{title}{Rising socio-economic costs of pm$_{2.5}$ pollution and medical service mismatching}}.
\newblock {\emph{\JournalTitle{Nature Sustainability}}} \bibinfo{pages}{1--11} (\bibinfo{year}{2025}).

\bibitem{3feng2024long}
\bibinfo{author}{Feng, Y.} \emph{et~al.}
\newblock \bibinfo{journal}{\bibinfo{title}{Long-term exposure to ambient pm$_{2.5}$, particulate constituents and hospital admissions from non-respiratory infection}}.
\newblock {\emph{\JournalTitle{Nature Communications}}} \textbf{\bibinfo{volume}{15}}, \bibinfo{pages}{1518} (\bibinfo{year}{2024}).

\bibitem{4epa2019isa}
\bibinfo{author}{{U.S. Environmental Protection Agency}}.
\newblock \bibinfo{title}{Integrated science assessment (isa) for particulate matter (final report, dec 2019)}.
\newblock \bibinfo{type}{Tech. Rep.} \bibinfo{number}{EPA/600/R-19/188}, \bibinfo{institution}{U.S. Environmental Protection Agency}, \bibinfo{address}{Washington, DC} (\bibinfo{year}{2019}).
\newblock \bibinfo{note}{Final Report}.

\bibitem{5liang2019urbanization}
\bibinfo{author}{Liang, L.}, \bibinfo{author}{Wang, Z.} \& \bibinfo{author}{Li, J.}
\newblock \bibinfo{journal}{\bibinfo{title}{The effect of urbanization on environmental pollution in rapidly developing urban agglomerations}}.
\newblock {\emph{\JournalTitle{Journal of Cleaner Production}}} \textbf{\bibinfo{volume}{237}}, \bibinfo{pages}{117649} (\bibinfo{year}{2019}).

\bibitem{6wang2021anthropogenic}
\bibinfo{author}{Wang, J.} \emph{et~al.}
\newblock \bibinfo{journal}{\bibinfo{title}{Anthropogenic emissions and urbanization increase risk of compound hot extremes in cities}}.
\newblock {\emph{\JournalTitle{Nature Climate Change}}} \textbf{\bibinfo{volume}{11}}, \bibinfo{pages}{1084--1089} (\bibinfo{year}{2021}).

\bibitem{7amann2011cost}
\bibinfo{author}{Amann, M.} \emph{et~al.}
\newblock \bibinfo{journal}{\bibinfo{title}{Cost-effective control of air quality and greenhouse gases in europe: Modeling and policy applications}}.
\newblock {\emph{\JournalTitle{Environmental Modelling \& Software}}} \textbf{\bibinfo{volume}{26}}, \bibinfo{pages}{1489--1501} (\bibinfo{year}{2011}).

\bibitem{8ekeh2025leveraging}
\bibinfo{author}{Ekeh, A.~H.}, \bibinfo{author}{Apeh, C.~E.}, \bibinfo{author}{Odionu, C.~S.} \& \bibinfo{author}{Austin-Gabriel, B.}
\newblock \bibinfo{journal}{\bibinfo{title}{Leveraging machine learning for environmental policy innovation: Advances in data analytics to address urban and ecological challenges}}.
\newblock {\emph{\JournalTitle{Gulf Journal of Advance Business Research}}} \textbf{\bibinfo{volume}{3}}, \bibinfo{pages}{456--482} (\bibinfo{year}{2025}).

\bibitem{9yang2023novel}
\bibinfo{author}{Yang, W.}, \bibinfo{author}{Wang, J.}, \bibinfo{author}{Zhang, K.} \& \bibinfo{author}{Hao, Y.}
\newblock \bibinfo{journal}{\bibinfo{title}{A novel air pollution forecasting, health effects, and economic cost assessment system for environmental management: From a new perspective of the district-level}}.
\newblock {\emph{\JournalTitle{Journal of Cleaner Production}}} \textbf{\bibinfo{volume}{417}}, \bibinfo{pages}{138027} (\bibinfo{year}{2023}).

\bibitem{10ding2022forecasting}
\bibinfo{author}{Ding, Z.}, \bibinfo{author}{Chen, H.}, \bibinfo{author}{Zhou, L.} \& \bibinfo{author}{Wang, Z.}
\newblock \bibinfo{journal}{\bibinfo{title}{A forecasting system for deterministic and uncertain prediction of air pollution data}}.
\newblock {\emph{\JournalTitle{Expert Systems with Applications}}} \textbf{\bibinfo{volume}{208}}, \bibinfo{pages}{118123} (\bibinfo{year}{2022}).

\bibitem{11liu2020exploring}
\bibinfo{author}{Liu, Y.}, \bibinfo{author}{Zhou, Y.} \& \bibinfo{author}{Lu, J.}
\newblock \bibinfo{journal}{\bibinfo{title}{Exploring the relationship between air pollution and meteorological conditions in china under environmental governance}}.
\newblock {\emph{\JournalTitle{Scientific Reports}}} \textbf{\bibinfo{volume}{10}}, \bibinfo{pages}{14518} (\bibinfo{year}{2020}).

\bibitem{12wen2024assessing}
\bibinfo{author}{Wen, W.} \emph{et~al.}
\newblock \bibinfo{journal}{\bibinfo{title}{Assessing aerosol-radiation interaction with wrf-chem-solar: Case study on the impact of the “pollution reduction and carbon reduction synergy” policy}}.
\newblock {\emph{\JournalTitle{Atmospheric Research}}} \textbf{\bibinfo{volume}{308}}, \bibinfo{pages}{107537} (\bibinfo{year}{2024}).

\bibitem{13mai2024convolutional}
\bibinfo{author}{Mai, Z.} \emph{et~al.}
\newblock \bibinfo{journal}{\bibinfo{title}{Convolutional neural networks facilitate process understanding of megacity ozone temporal variability}}.
\newblock {\emph{\JournalTitle{Environmental Science \& Technology}}} \textbf{\bibinfo{volume}{58}}, \bibinfo{pages}{15691--15701} (\bibinfo{year}{2024}).

\bibitem{14zhou2024impacts}
\bibinfo{author}{Zhou, M.}, \bibinfo{author}{Xie, Y.}, \bibinfo{author}{Wang, C.}, \bibinfo{author}{Shen, L.} \& \bibinfo{author}{Mauzerall, D.~L.}
\newblock \bibinfo{journal}{\bibinfo{title}{Impacts of current and climate induced changes in atmospheric stagnation on indian surface pm$_{2.5}$ pollution}}.
\newblock {\emph{\JournalTitle{Nature Communications}}} \textbf{\bibinfo{volume}{15}}, \bibinfo{pages}{7448} (\bibinfo{year}{2024}).

\bibitem{15han2020local}
\bibinfo{author}{Han, H.}, \bibinfo{author}{Liu, J.}, \bibinfo{author}{Shu, L.}, \bibinfo{author}{Wang, T.} \& \bibinfo{author}{Yuan, H.}
\newblock \bibinfo{journal}{\bibinfo{title}{Local and synoptic meteorological influences on daily variability in summertime surface ozone in eastern china}}.
\newblock {\emph{\JournalTitle{Atmospheric Chemistry and Physics}}} \textbf{\bibinfo{volume}{20}}, \bibinfo{pages}{203--222} (\bibinfo{year}{2020}).

\bibitem{16schnell2018exploring}
\bibinfo{author}{Schnell, J.~L.} \emph{et~al.}
\newblock \bibinfo{journal}{\bibinfo{title}{Exploring the relationship between surface pm$_{2.5}$ and meteorology in northern india}}.
\newblock {\emph{\JournalTitle{Atmospheric Chemistry and Physics}}} \textbf{\bibinfo{volume}{18}}, \bibinfo{pages}{10157--10175} (\bibinfo{year}{2018}).

\bibitem{17yuval2020association}
\bibinfo{author}{Yuval, Y.~L.}, \bibinfo{author}{Dayan, U.}, \bibinfo{author}{Levy, I.} \& \bibinfo{author}{Broday, D.~M.}
\newblock \bibinfo{journal}{\bibinfo{title}{On the association between characteristics of the atmospheric boundary layer and air pollution concentrations}}.
\newblock {\emph{\JournalTitle{Atmospheric Research}}} \textbf{\bibinfo{volume}{231}}, \bibinfo{pages}{104675} (\bibinfo{year}{2020}).

\bibitem{18miao2021relationship}
\bibinfo{author}{Miao, Y.}, \bibinfo{author}{Che, H.}, \bibinfo{author}{Zhang, X.} \& \bibinfo{author}{Liu, S.}
\newblock \bibinfo{journal}{\bibinfo{title}{Relationship between summertime concurring pm$_{2.5}$ and o$_{3}$ pollution and boundary layer height differs between beijing and shanghai, china}}.
\newblock {\emph{\JournalTitle{Environmental Pollution}}} \textbf{\bibinfo{volume}{268}}, \bibinfo{pages}{115775} (\bibinfo{year}{2021}).

\bibitem{19zhang2023evolution}
\bibinfo{author}{Zhang, Y.} \emph{et~al.}
\newblock \bibinfo{journal}{\bibinfo{title}{Evolution of ozone formation sensitivity during a persistent regional ozone episode in northeastern china and its implication for a control strategy}}.
\newblock {\emph{\JournalTitle{Environmental Science \& Technology}}} \textbf{\bibinfo{volume}{58}}, \bibinfo{pages}{617--627} (\bibinfo{year}{2023}).

\bibitem{20sulaymon2023using}
\bibinfo{author}{Sulaymon, I.~D.} \emph{et~al.}
\newblock \bibinfo{journal}{\bibinfo{title}{Using the covid-19 lockdown to identify atmospheric processes and meteorology influences on regional pm$_{2.5}$ pollution episodes in the beijing-tianjin-hebei, china}}.
\newblock {\emph{\JournalTitle{Atmospheric Research}}} \textbf{\bibinfo{volume}{294}}, \bibinfo{pages}{106940} (\bibinfo{year}{2023}).

\bibitem{21tie2010impact}
\bibinfo{author}{Tie, X.}, \bibinfo{author}{Brasseur, G.} \& \bibinfo{author}{Ying, Z.}
\newblock \bibinfo{journal}{\bibinfo{title}{Impact of model resolution on chemical ozone formation in mexico city: application of the wrf-chem model}}.
\newblock {\emph{\JournalTitle{Atmospheric Chemistry and Physics}}} \textbf{\bibinfo{volume}{10}}, \bibinfo{pages}{8983--8995} (\bibinfo{year}{2010}).

\bibitem{22wang2025causal}
\bibinfo{author}{Wang, L.} \emph{et~al.}
\newblock \bibinfo{journal}{\bibinfo{title}{Causal-inference machine learning reveals the drivers of china's 2022 ozone rebound}}.
\newblock {\emph{\JournalTitle{Environmental Science and Ecotechnology}}} \textbf{\bibinfo{volume}{24}}, \bibinfo{pages}{100524} (\bibinfo{year}{2025}).

\bibitem{23li2019ozone}
\bibinfo{author}{Li, L.} \emph{et~al.}
\newblock \bibinfo{journal}{\bibinfo{title}{Ozone source apportionment over the yangtze river delta region, china: Investigation of regional transport, sectoral contributions and seasonal differences}}.
\newblock {\emph{\JournalTitle{Atmospheric Environment}}} \textbf{\bibinfo{volume}{202}}, \bibinfo{pages}{269--280} (\bibinfo{year}{2019}).

\bibitem{24zhang2020source}
\bibinfo{author}{Zhang, C.} \emph{et~al.}
\newblock \bibinfo{journal}{\bibinfo{title}{Source assessment of atmospheric fine particulate matter in a chinese megacity: Insights from long-term, high-time resolution chemical composition measurements from shanghai flagship monitoring supersite}}.
\newblock {\emph{\JournalTitle{Chemosphere}}} \textbf{\bibinfo{volume}{251}}, \bibinfo{pages}{126598} (\bibinfo{year}{2020}).

\bibitem{25yu2010eta}
\bibinfo{author}{Yu, S.} \emph{et~al.}
\newblock \bibinfo{journal}{\bibinfo{title}{Eta-cmaq air quality forecasts for o$_3$ and related species using three different photochemical mechanisms (cb4, cb05, saprc-99): comparisons with measurements during the 2004 icartt study}}.
\newblock {\emph{\JournalTitle{Atmospheric Chemistry and Physics}}} \textbf{\bibinfo{volume}{10}}, \bibinfo{pages}{3001--3025} (\bibinfo{year}{2010}).

\bibitem{27chen2023fuxi}
\bibinfo{author}{Chen, L.} \emph{et~al.}
\newblock \bibinfo{journal}{\bibinfo{title}{Fuxi: A cascade machine learning forecasting system for 15-day global weather forecast}}.
\newblock {\emph{\JournalTitle{npj climate and atmospheric science}}} \textbf{\bibinfo{volume}{6}}, \bibinfo{pages}{190} (\bibinfo{year}{2023}).

\bibitem{28lam2023learning}
\bibinfo{author}{Lam, R.} \emph{et~al.}
\newblock \bibinfo{journal}{\bibinfo{title}{Learning skillful medium-range global weather forecasting}}.
\newblock {\emph{\JournalTitle{Science}}} \textbf{\bibinfo{volume}{382}}, \bibinfo{pages}{1416--1421} (\bibinfo{year}{2023}).

\bibitem{29pathak2022fourcastnet}
\bibinfo{author}{Pathak, J.} \emph{et~al.}
\newblock \bibinfo{journal}{\bibinfo{title}{Fourcastnet: A global data-driven high-resolution weather model using adaptive fourier neural operators}}.
\newblock {\emph{\JournalTitle{arXiv preprint arXiv:2202.11214}}}  (\bibinfo{year}{2022}).

\bibitem{wu2023hybrid}
\bibinfo{author}{Wu, C.-l.} \emph{et~al.}
\newblock \bibinfo{journal}{\bibinfo{title}{A hybrid deep learning model for regional o$_{3}$ and no$_{2}$ concentrations prediction based on spatiotemporal dependencies in air quality monitoring network}}.
\newblock {\emph{\JournalTitle{Environmental pollution}}} \textbf{\bibinfo{volume}{320}}, \bibinfo{pages}{121075} (\bibinfo{year}{2023}).

\bibitem{cheng2021development}
\bibinfo{author}{Cheng, Y.}, \bibinfo{author}{He, L.-Y.} \& \bibinfo{author}{Huang, X.-F.}
\newblock \bibinfo{journal}{\bibinfo{title}{Development of a high-performance machine learning model to predict ground ozone pollution in typical cities of china}}.
\newblock {\emph{\JournalTitle{Journal of Environmental Management}}} \textbf{\bibinfo{volume}{299}}, \bibinfo{pages}{113670} (\bibinfo{year}{2021}).

\bibitem{qi2019hybrid}
\bibinfo{author}{Qi, Y.}, \bibinfo{author}{Li, Q.}, \bibinfo{author}{Karimian, H.} \& \bibinfo{author}{Liu, D.}
\newblock \bibinfo{journal}{\bibinfo{title}{A hybrid model for spatiotemporal forecasting of pm$_{2.5}$ based on graph convolutional neural network and long short-term memory}}.
\newblock {\emph{\JournalTitle{Science of The Total Environment}}} \textbf{\bibinfo{volume}{664}}, \bibinfo{pages}{1--10} (\bibinfo{year}{2019}).

\bibitem{bodnar2025foundation}
\bibinfo{author}{Bodnar, C.} \emph{et~al.}
\newblock \bibinfo{journal}{\bibinfo{title}{A foundation model for the earth system}}.
\newblock {\emph{\JournalTitle{Nature}}}  (\bibinfo{year}{2025}).

\bibitem{30zhu2024hybrid}
\bibinfo{author}{Zhu, J.}, \bibinfo{author}{Niu, L.}, \bibinfo{author}{Zheng, P.}, \bibinfo{author}{Chen, H.} \& \bibinfo{author}{Liu, J.}
\newblock \bibinfo{journal}{\bibinfo{title}{A hybrid pm$_{2.5}$ interval concentration prediction framework based on multi-factor interval decomposition reconstruction strategy and attention mechanism}}.
\newblock {\emph{\JournalTitle{Atmospheric Environment}}} \textbf{\bibinfo{volume}{335}}, \bibinfo{pages}{120730} (\bibinfo{year}{2024}).

\bibitem{31mandal2023city}
\bibinfo{author}{Mandal, S.} \& \bibinfo{author}{Thakur, M.}
\newblock \bibinfo{journal}{\bibinfo{title}{A city-based pm$_{2.5}$ forecasting framework using spatially attentive cluster-based graph neural network model}}.
\newblock {\emph{\JournalTitle{Journal of Cleaner Production}}} \textbf{\bibinfo{volume}{405}}, \bibinfo{pages}{137036} (\bibinfo{year}{2023}).

\bibitem{32zhang2024long}
\bibinfo{author}{Zhang, C.}, \bibinfo{author}{Wang, S.}, \bibinfo{author}{Wu, Y.}, \bibinfo{author}{Zhu, X.} \& \bibinfo{author}{Shen, W.}
\newblock \bibinfo{journal}{\bibinfo{title}{A long-term prediction method for pm$_{2.5}$ concentration based on spatiotemporal graph attention recurrent neural network and grey wolf optimization algorithm}}.
\newblock {\emph{\JournalTitle{Journal of Environmental Chemical Engineering}}} \textbf{\bibinfo{volume}{12}}, \bibinfo{pages}{111716} (\bibinfo{year}{2024}).

\bibitem{33liu2022prediction}
\bibinfo{author}{Liu, H.}
\newblock \bibinfo{title}{Prediction of air pollutant concentration based on self-attention mechanism lstm model}.
\newblock In \emph{\bibinfo{booktitle}{International Conference on High Performance Computing and Communication (HPCCE 2021)}}, vol. \bibinfo{volume}{12162}, \bibinfo{pages}{325--329} (\bibinfo{organization}{SPIE}, \bibinfo{year}{2022}).

\bibitem{34guyu2025pm2}
\bibinfo{author}{Guyu, Z.}, \bibinfo{author}{Xiaoyuan, Y.}, \bibinfo{author}{Jiansen, S.}, \bibinfo{author}{Hongdou, H.} \& \bibinfo{author}{Qian, W.}
\newblock \bibinfo{journal}{\bibinfo{title}{A pm$_{2.5}$ spatiotemporal prediction model based on mixed graph convolutional gru and self-attention network}}.
\newblock {\emph{\JournalTitle{Environmental Pollution}}} \bibinfo{pages}{125748} (\bibinfo{year}{2025}).

\bibitem{35zhang2023air}
\bibinfo{author}{Zhang, B.} \emph{et~al.}
\newblock \bibinfo{journal}{\bibinfo{title}{Air pollutant diffusion trend prediction based on deep learning for targeted season—north china as an example}}.
\newblock {\emph{\JournalTitle{Expert Systems with Applications}}} \textbf{\bibinfo{volume}{232}}, \bibinfo{pages}{120718} (\bibinfo{year}{2023}).

\bibitem{36wang2022air}
\bibinfo{author}{Wang, Z.}, \bibinfo{author}{Yang, Y.} \& \bibinfo{author}{Yue, S.}
\newblock \bibinfo{journal}{\bibinfo{title}{Air quality classification and measurement based on double output vision transformer}}.
\newblock {\emph{\JournalTitle{IEEE Internet of Things Journal}}} \textbf{\bibinfo{volume}{9}}, \bibinfo{pages}{20975--20984} (\bibinfo{year}{2022}).

\bibitem{37baruah2024novel}
\bibinfo{author}{Baruah, A.} \emph{et~al.}
\newblock \bibinfo{journal}{\bibinfo{title}{A novel spatiotemporal prediction approach to fill air pollution data gaps using mobile sensors, machine learning and citizen science techniques}}.
\newblock {\emph{\JournalTitle{npj Climate and Atmospheric Science}}} \textbf{\bibinfo{volume}{7}}, \bibinfo{pages}{310} (\bibinfo{year}{2024}).

\bibitem{38li2024multi}
\bibinfo{author}{Li, B.} \& \bibinfo{author}{Wang, P.}
\newblock \bibinfo{journal}{\bibinfo{title}{A multi-task stations cooperative air quality prediction system for sustainable development}}.
\newblock {\emph{\JournalTitle{Humanities and Social Sciences Communications}}} \textbf{\bibinfo{volume}{11}}, \bibinfo{pages}{1--11} (\bibinfo{year}{2024}).

\bibitem{39liu2024air}
\bibinfo{author}{Liu, B.}, \bibinfo{author}{Lai, M.}, \bibinfo{author}{Zeng, P.} \& \bibinfo{author}{Chen, J.}
\newblock \bibinfo{journal}{\bibinfo{title}{Air pollutant prediction based on a attention mechanism model of the yangtze river delta region in frequent heatwaves}}.
\newblock {\emph{\JournalTitle{Atmospheric Research}}} \textbf{\bibinfo{volume}{311}}, \bibinfo{pages}{107701} (\bibinfo{year}{2024}).

\bibitem{zhang2016air}
\bibinfo{author}{Zhang, H.} \emph{et~al.}
\newblock \bibinfo{journal}{\bibinfo{title}{Air pollution and control action in beijing}}.
\newblock {\emph{\JournalTitle{Journal of Cleaner Production}}} \textbf{\bibinfo{volume}{112}}, \bibinfo{pages}{1519--1527} (\bibinfo{year}{2016}).

\bibitem{feng2014formation}
\bibinfo{author}{Feng, X.} \emph{et~al.}
\newblock \bibinfo{journal}{\bibinfo{title}{Formation and dominant factors of haze pollution over beijing and its peripheral areas in winter}}.
\newblock {\emph{\JournalTitle{Atmospheric Pollution Research}}} \textbf{\bibinfo{volume}{5}}, \bibinfo{pages}{528--538} (\bibinfo{year}{2014}).

\bibitem{chang2021meteorology}
\bibinfo{author}{Chang, L.}, \bibinfo{author}{He, F.}, \bibinfo{author}{Tie, X.}, \bibinfo{author}{Xu, J.} \& \bibinfo{author}{Gao, W.}
\newblock \bibinfo{journal}{\bibinfo{title}{Meteorology driving the highest ozone level occurred during mid-spring to early summer in shanghai, china}}.
\newblock {\emph{\JournalTitle{Science of the Total Environment}}} \textbf{\bibinfo{volume}{785}}, \bibinfo{pages}{147253} (\bibinfo{year}{2021}).

\bibitem{coates2016influence}
\bibinfo{author}{Coates, J.}, \bibinfo{author}{Mar, K.~A.}, \bibinfo{author}{Ojha, N.} \& \bibinfo{author}{Butler, T.~M.}
\newblock \bibinfo{journal}{\bibinfo{title}{The influence of temperature on ozone production under varying no$_{x}$ conditions--a modelling study}}.
\newblock {\emph{\JournalTitle{Atmospheric Chemistry and Physics}}} \textbf{\bibinfo{volume}{16}}, \bibinfo{pages}{11601--11615} (\bibinfo{year}{2016}).

\bibitem{seinfeld2016atmospheric}
\bibinfo{author}{Seinfeld, J.~H.} \& \bibinfo{author}{Pandis, S.~N.}
\newblock \emph{\bibinfo{title}{Atmospheric chemistry and physics: from air pollution to climate change}} (\bibinfo{publisher}{John Wiley \& Sons}, \bibinfo{year}{2016}).

\bibitem{kong2016empirical}
\bibinfo{author}{Kong, L.}, \bibinfo{author}{Xin, J.}, \bibinfo{author}{Zhang, W.} \& \bibinfo{author}{Wang, Y.}
\newblock \bibinfo{journal}{\bibinfo{title}{The empirical correlations between pm$_{2.5}$, pm$_{10}$ and aod in the beijing metropolitan region and the pm$_{2.5}$, pm$_{10}$ distributions retrieved by modis}}.
\newblock {\emph{\JournalTitle{Environmental pollution}}} \textbf{\bibinfo{volume}{216}}, \bibinfo{pages}{350--360} (\bibinfo{year}{2016}).

\bibitem{li2016observation}
\bibinfo{author}{Li, Y.} \emph{et~al.}
\newblock \bibinfo{journal}{\bibinfo{title}{Observation of regional air pollutant transport between the megacity beijing and the north china plain}}.
\newblock {\emph{\JournalTitle{Atmospheric Chemistry and Physics}}} \textbf{\bibinfo{volume}{16}}, \bibinfo{pages}{14265--14283} (\bibinfo{year}{2016}).

\bibitem{yin2025regional}
\bibinfo{author}{Yin, L.} \emph{et~al.}
\newblock \bibinfo{journal}{\bibinfo{title}{Regional-specific trends of pm2. 5 and o3 temperature sensitivity in the united states}}.
\newblock {\emph{\JournalTitle{npj Climate and Atmospheric Science}}} \textbf{\bibinfo{volume}{8}}, \bibinfo{pages}{12} (\bibinfo{year}{2025}).

\bibitem{granier2019copernicus}
\bibinfo{author}{Granier, C.} \emph{et~al.}
\newblock \emph{\bibinfo{title}{The Copernicus atmosphere monitoring service global and regional emissions (April 2019 version)}}.
\newblock Ph.D. thesis, \bibinfo{school}{Copernicus Atmosphere Monitoring Service} (\bibinfo{year}{2019}).

\bibitem{soulie2023global}
\bibinfo{author}{Soulie, A.} \emph{et~al.}
\newblock \bibinfo{journal}{\bibinfo{title}{Global anthropogenic emissions (cams-glob-ant) for the copernicus atmosphere monitoring service simulations of air quality forecasts and reanalyses}}.
\newblock {\emph{\JournalTitle{Earth System Science Data Discussions}}} \textbf{\bibinfo{volume}{2023}}, \bibinfo{pages}{1--45} (\bibinfo{year}{2023}).

\bibitem{hersbach2020era5}
\bibinfo{author}{Hersbach, H.} \emph{et~al.}
\newblock \bibinfo{journal}{\bibinfo{title}{The era5 global reanalysis}}.
\newblock {\emph{\JournalTitle{Quarterly journal of the royal meteorological society}}} \textbf{\bibinfo{volume}{146}}, \bibinfo{pages}{1999--2049} (\bibinfo{year}{2020}).

\bibitem{zhong2024fuxi}
\bibinfo{author}{Zhong, X.} \emph{et~al.}
\newblock \bibinfo{journal}{\bibinfo{title}{Fuxi-2.0: Advancing machine learning weather forecasting model for practical applications}}.
\newblock {\emph{\JournalTitle{arXiv preprint arXiv:2409.07188}}}  (\bibinfo{year}{2024}).

\bibitem{sun2024fuxi}
\bibinfo{author}{Sun, X.} \emph{et~al.}
\newblock \bibinfo{journal}{\bibinfo{title}{Fuxi weather: An end-to-end machine learning weather data assimilation and forecasting system}}.
\newblock {\emph{\JournalTitle{arXiv preprint arXiv:2408.05472}}}  (\bibinfo{year}{2024}).

\end{thebibliography}

\section*{Supplementary materials}

\subsection*{Supplementary materials A. Datasets Introduction}
\begin{itemize}[label=$\blacklozenge$] 
 \item CAMS-GLOB-ANT: CAMS-GLOB-ANT (Copernicus Atmosphere Monitoring Service Global Anthropogenic Emissions) \cite{granier2019copernicus,soulie2023global} is a high-resolution global anthropogenic emissions inventory developed by the Copernicus Atmosphere Monitoring Service (CAMS) of the European Union. It is designed to provide dynamic emission data support for air quality forecasting, climate modeling, and pollution source attribution. This dataset covers 36 atmospheric constituents (e.g., NO$_x$, SO$_{2}$, CO, PM$_{2.5}$, CO$_{2}$, etc.) across 17 economic sectors (e.g., transportation, energy, industry), with monthly averaged emission values available for the period from 2000 to 2023. The spatial resolution is 0.1° × 0.1°. In this study, CAMS-GLOB-ANT data from 2016 to 2023 were used.
 \item ERA5: ERA5 (ECMWF ReAnalysis 5th Generation) is the fifth-generation global atmospheric reanalysis dataset released by the European Centre for Medium-Range Weather Forecasts (ECMWF) \cite{hersbach2020era5}. It is a high-precision reanalysis product generated by combining multi-source observational data with numerical simulations, and is widely used for both weather and climate research. The dataset spans from January 1950 to the present, with a temporal resolution of 1 hour and a spatial resolution of 0.25° × 0.25°. To better meet the requirements of this study, bilinear interpolation was applied to downscale the original data to a spatial resolution of 0.1° × 0.1°. ERA5 is known for its high reliability in representing meteorological variables and has been extensively applied in research related to atmospheric environment and climate change.
 \item FuXi-2.0: FuXi-2.0 is a weather forecasting model developed based on the ERA5 reanalysis dataset from the European Centre for Medium-Range Weather Forecasts (ECMWF)\cite{zhong2024fuxi,sun2024fuxi} . It provides global forecasts at an hourly temporal resolution and a spatial resolution of 0.25° × 0.25°, covering a comprehensive set of meteorological variables. FuXi-2.0 outperforms traditional models such as ECMWF HRES in applications like wind energy forecasting and tropical cyclone intensity prediction, while requiring significantly fewer computational resources [4,5]. In this study, FuXi forecasts initialized at 00:00 and 12:00 Coordinated Universal Time (UTC) were used as meteorological inputs during the model testing phase. These forecasts have a 1-hour temporal resolution and were downscaled to 0.1° × 0.1° using the same bilinear interpolation method as applied to ERA5. The forecast horizon was set to 72 hours. 
 \item Station Pollutant Data: This study collected continuous air quality monitoring data from national control (Guokong) stations in Beijing, Shanghai, and Shenzhen spanning the period from 2016 to 2023. The monitored indicators include conventional gaseous pollutants—O$_{3}$, NO$_{2}$, PM$_{2.5}$, PM$_{10}$, SO$_{2}$—as well as the Air Quality Index (AQI). All data were obtained from the China National Environmental Monitoring Center (CNEMC) through the national air quality monitoring network and underwent strict quality control procedures, including validation checks, outlier removal, and missing data treatment. After filtering, valid monitoring data from 11 stations in Beijing, 19 in Shanghai, and 11 in Shenzhen were retained for analysis. All stations are operated and maintained in accordance with the “Technical Specifications for Ambient Air Quality Monitoring” (HJ 664-2013), ensuring the accuracy and comparability of the data.
 \item Detailed dataset information is provided in Supplementary Tables 1 and 2. 
\end{itemize}

\renewcommand{\thetable}{\arabic{table}} 
\renewcommand{\tablename}{Supplementary Table} 
\setcounter{table}{0} 

\renewcommand{\thefigure}{\arabic{figure}}  
\captionsetup[figure]{labelfont=bf, labelsep=period, name=Supplementary Figure}

\setcounter{figure}{0}

\begin{table}[H]
    \centering
    \scriptsize
    \renewcommand{\arraystretch}{1.3} 
    \setlength{\tabcolsep}{1.2pt}  
    \caption{Summary of the datasets used for model training and evaluation in this study.}
    \label{tab:Supplementary Table 1}
    
\begin{tabular}{ccccc}
\hline
Name                   & Resolution & Timeframe & Surface Variables                      & \begin{tabular}[c]{@{}c@{}}Pressure\\ Variables\end{tabular} \\ \hline
CAMS                   & 0.1°×0.1°  & 2016-2023 & BC, OC, SO$_{2}$, N2O, NH3, NMVOCs, NOx, CO & –                                                            \\
ERA5                   & 0.1°×0.1°  & 2016-2023 & T2M, D2M, U10M, V10M, V100M, U100M, TP & U, V, T, SH                                                  \\
FuXi                   & 0.1°×0.1°  & 2019-2023 & T2M, D2M, U10M, V10M, V100M, U100M, TP & U, V, T, SH                                                  \\
Station Pollutant Data & –          & 2016-2023 & AQI, O$_{3}$, NO$_{2}$, PM$_{2.5}$, PM$_{10}$, SO$_{2}$, CO     & –                                                            \\
Geographical           & 0.1°×0.1°  & –         & land-seamask, latitude, longitude      & –                                                            \\
Temporal               & –          & –         & hour of day, day of year, step         & –                                                            \\ \hline
\end{tabular}
\end{table}

\begin{table}[H]
    \centering
    \scriptsize
    \renewcommand{\arraystretch}{1.3} 
    \setlength{\tabcolsep}{1.2pt}  
    \caption{Statistical summary of five ablation experiment groups in three cities.}
    \label{tab:Supplementary Table 2}
\begin{tabular}{cccccc}
\hline
\multirow{2}{*}{City}     & \multirow{2}{*}{Exp. ID} & \multirow{2}{*}{Description}                                                                                                     & \multirow{2}{*}{Data Types}                                                        & \multicolumn{2}{c}{Sample Size} \\
                          &                          &                                                                                                                                  &                                                                                    & Train           & Test          \\ \hline
\multirow{4}{*}{Beijing}  & A1                       & \begin{tabular}[c]{@{}c@{}}All data included:\\ emission inventory,\\ meteorology,\\ Station pollutant data\\ (ALL)\end{tabular} & \begin{tabular}[c]{@{}c@{}}CAMS, ERA5, FuXi,\\ Station pollutant data\end{tabular} & 3337            & 709           \\
                          & B1                       & \begin{tabular}[c]{@{}c@{}}Excluding meteorological data\\ (DEMET)\end{tabular}                                                  & \begin{tabular}[c]{@{}c@{}}CAMS,\\ Station pollutant data\end{tabular}             & 3337            & 709           \\
                          & C1                       & \begin{tabular}[c]{@{}c@{}}Excluding emission inventory data\\ (DEEMS)\end{tabular}                                              & \begin{tabular}[c]{@{}c@{}}ERA5, FuXi,\\ Station pollutant data\end{tabular}       & 3337            & 709           \\
                          & D1                       & \begin{tabular}[c]{@{}c@{}}Only Station pollutant data\\ (STN\_ONLY)\end{tabular}                                                & Station pollutant data                                                             & 3337            & 709           \\
\multirow{4}{*}{Shanghai} & A2                       & \begin{tabular}[c]{@{}c@{}}All data included\\ (ALL)\end{tabular}                                                                & \begin{tabular}[c]{@{}c@{}}CAMS, ERA5, FuXi,\\ Station pollutant data\end{tabular} & 7619            & 709           \\
                          & B2                       & \begin{tabular}[c]{@{}c@{}}Excluding meteorological data\\ (DEMET)\end{tabular}                                                  & \begin{tabular}[c]{@{}c@{}}CAMS,\\ Station pollutant data\end{tabular}             & 7619            & 709           \\
                          & C2                       & \begin{tabular}[c]{@{}c@{}}Excluding emission inventory data\\ (DEEMS)\end{tabular}                                              & \begin{tabular}[c]{@{}c@{}}ERA5, FuXi,\\ Station pollutant data\end{tabular}       & 7619            & 709           \\
                          & D2                       & \begin{tabular}[c]{@{}c@{}}Only Station pollutant data\\ (STN\_ONLY)\end{tabular}                                                & Station pollutant data                                                             & 7619            & 709           \\
\multirow{4}{*}{Shenzhen} & A3                       & All data included                                                                                                                & \begin{tabular}[c]{@{}c@{}}CAMS, ERA5, FuXi,\\ Station pollutant data\end{tabular} & 6060            & 709           \\
                          & B3                       & \begin{tabular}[c]{@{}c@{}}Excluding meteorological data\\ (DEMET)\end{tabular}                                                  & CAMS, Station pollutant data                                                       & 6060            & 709           \\
                          & C3                       & \begin{tabular}[c]{@{}c@{}}Excluding emission inventory data\\ (DEEMS)\end{tabular}                                              & \begin{tabular}[c]{@{}c@{}}ERA5, FuXi,\\ Station pollutant data\end{tabular}       & 6060            & 709           \\
                          & D3                       & \begin{tabular}[c]{@{}c@{}}Only Station pollutant data\\ (STN\_ONLY)\end{tabular}                                                & Station pollutant data                                                             & 6060            & 709           \\ \hline
\end{tabular}
\end{table}

\subsection*{Supplementary materials B. Model Architecture and Evaluation Metrics}
\begin{itemize}[label=$\blacklozenge$]  
  \item Pollutant Monitoring Site Interaction Module 

The implementation details are as follows:

\begin{equation}
\text{PE}_{ij}^{\text{lat}} =
\left[
\frac{\sin(\text{lat}_{ij} - \text{lat}_{0,o})}{\text{lat}_{\text{range}}},
\frac{\cos(\text{lat}_{ij} - \text{lat}_{0,o})}{\text{lat}_{\text{range}}}
\right]\\
\end{equation}
\begin{align}
T_{\text{emb}} = \text{Embedding}(t_{\text{doy}}, t_{\text{hod}})
\end{align}
\begin{align}
X_{\text{inp}} = \text{concat}(X_{t-6}, \text{PE}^{\text{lat,lon}}, T_{\text{emb}})
\end{align}
\begin{align}
X_{\text{sa}} = \text{SelfAttention}(X_{\text{inp}})
\end{align}

Where: \( X_{\text{inp}} \in \mathbb{R}^{(2D) \times N_s} \), \( D \) denotes the dimensionality of pollutant features at each site, \( N_s \) represents the number of monitoring sites, PE refers to the relative positional encoding derived from site longitude and latitude, \( T_{\text{emb}} \) denotes the embedded temporal encoding.

The detailed self-attention mechanism for site interaction is described as follows:

\begin{align}
Q, K, V &= W^q \cdot X_{\text{inp}},\quad W^k \cdot X_{\text{inp}},\quad W^v \cdot X_{\text{inp}} \\
A &= \text{softmax} \left( \frac{QK^\top}{\sqrt{d_{N_s}}} \right) \\
H_{\text{sa}} &= AV \\
H'_{\text{sa}} &= \text{MLP}(\text{LayerNorm}(V + H_{\text{sa}}))
\end{align}

\( A \in \mathbb{R}^{N_s \times N_s} \), the self-attention matrix \( A \) quantifies the relative importance among different monitoring sites. By multiplying \( A \) with the value matrix \( V \), we obtain the interaction-enhanced latent feature representations for each site. Subsequently, a feedforward layer is applied to perform nonlinear transformation and feature extraction over the residual structure, which enhances the model’s representational capacity and mitigates the risk of overfitting.

  \item Meteorology and Emission Inventory Coupling Module 

The implementation details are as follows:

\begin{align}
H_{\text{EM}} = \text{ResNet} \left( \text{concat}( \text{Met}_{\text{inp}}, \text{EMS}_{\text{inp}}, \text{PE}^{\text{lon,lat}} ) \right)
\end{align}

\begin{align}
Q, K, V &= W^q \cdot H'_{\text{sa}},\quad W^k \cdot H_{\text{EM}},\quad W^v \cdot H_{\text{EM}} \\
A_{\text{cross}} &= \text{softmax} \left( \frac{QK^\top}{\sqrt{d_{N_s}}} \right) \\
H_{\text{MEA}} &= A_{\text{cross}} V \\
H'_{\text{MEA}} &= \text{MLP} \left( \text{LayerNorm}(Q + H_{\text{MEA}}) \right)
\end{align}

\noindent
\( A_{\text{cross}} \in \mathbb{R}^{N_s \times N_{\text{grids}}} \). The matrix \( A \) denotes the cross-attention matrix, while \( H_{\text{EM}} \) represents the latent features extracted from meteorological and emission inventory inputs. \( N_{\text{grids}} \) refers to the number of spatial grid points in \( H_{\text{EM}} \). The cross-attention matrix \( A_{\text{cross}} \) captures the interaction between site-level latent variables and \( H_{\text{EM}} \), and its product with the value matrix \( V \) yields the coupled latent feature representation.A feedforward network is then applied to the residual structure for nonlinear transformation and feature extraction. Since meteorological and emission inventories primarily serve as contextual background information, the residual connection is constructed by adding the Query (Q) to the attention-enhanced representation , allowing for effective fusion of the cross-attention features while preserving the integrity of the input information.

  \item To comprehensively evaluate the forecasting performance of the model for six major pollutants across Beijing, Shanghai, and Shenzhen in 2023, five widely used statistical indicators are employed. These include the Root Mean Square Error (RMSE), Mean Absolute Error (MAE), Mean Relative Error (MRE), Relative Root Mean Square Error (rRMSE), and the Pearson correlation coefficient (R). The mathematical definitions of these indicators are as follows:

\begin{equation}
R = \frac{ \sum_{i=1}^{N} (P_i - \bar{P})(O_i - \bar{O}) }{ \sqrt{ \sum_{i=1}^{N} (P_i - \bar{P})^2 } \sqrt{ \sum_{i=1}^{N} (O_i - \bar{O})^2 } }
\end{equation}

\begin{equation}
\text{RMSE} = \sqrt{ \frac{1}{N} \sum_{i=1}^{N} (P_i - O_i)^2 }
\end{equation}

\begin{equation}
\text{rRMSE} = \frac{ \text{RMSE} }{ \bar{O} } = \frac{ \sqrt{ \frac{1}{N} \sum_{i=1}^{N} (P_i - O_i)^2 } }{ \bar{O} }
\end{equation}

\begin{equation}
\text{MRE} = \frac{1}{N} \sum_{i=1}^{N} \left| \frac{P_i - O_i}{O_i} \right|
\end{equation}

\begin{equation}
\text{MAE} = \frac{1}{N} \sum_{i=1}^{N} |P_i - O_i|
\end{equation}

where ${P}_i$ denotes the predicted concentration of the target pollutant; ${O}_i$  denotes the observed concentration; $\bar{P}$ and $\bar{O}$ represent the mean values of the predicted and observed concentrations, respectively; and N is the total number of samples.

\end{itemize}

\subsection*{Supplementary materials C. Figures and Tables}

\begin{figure}[H]
\centering
\includegraphics[height=0.7\textheight]{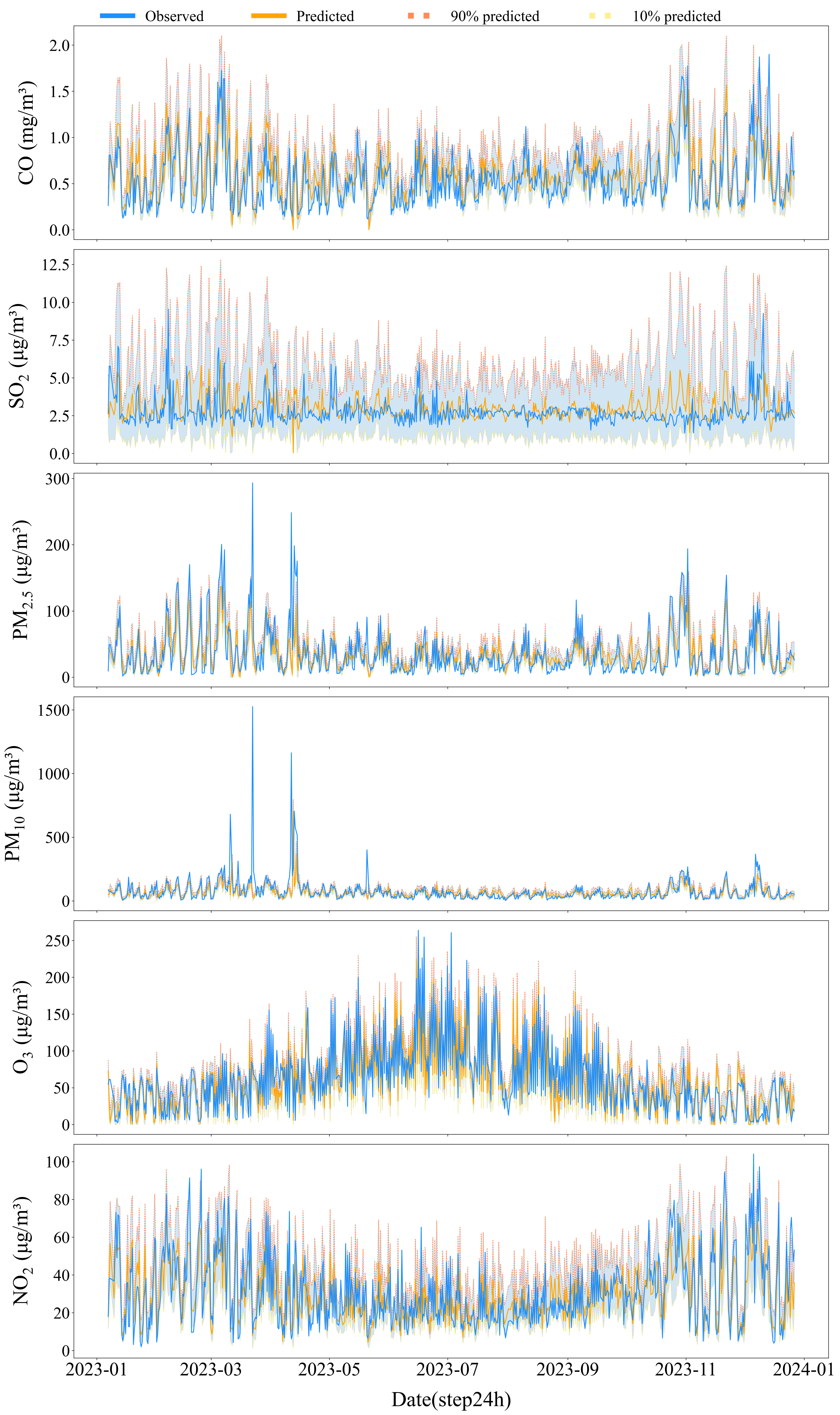}
\caption{Comparison of simulated and observed concentrations of six pollutants in beijing at 24-hour. “Forecast” refers to the point forecast, “90\% forecast” and “10\% forecast” represent the 90th and 10th percentile forecasts, respectively.}
\label{Supplementary Fig:beijing_24h}
\end{figure}

\begin{figure}[H]
\centering
\includegraphics[height=0.7\textheight]{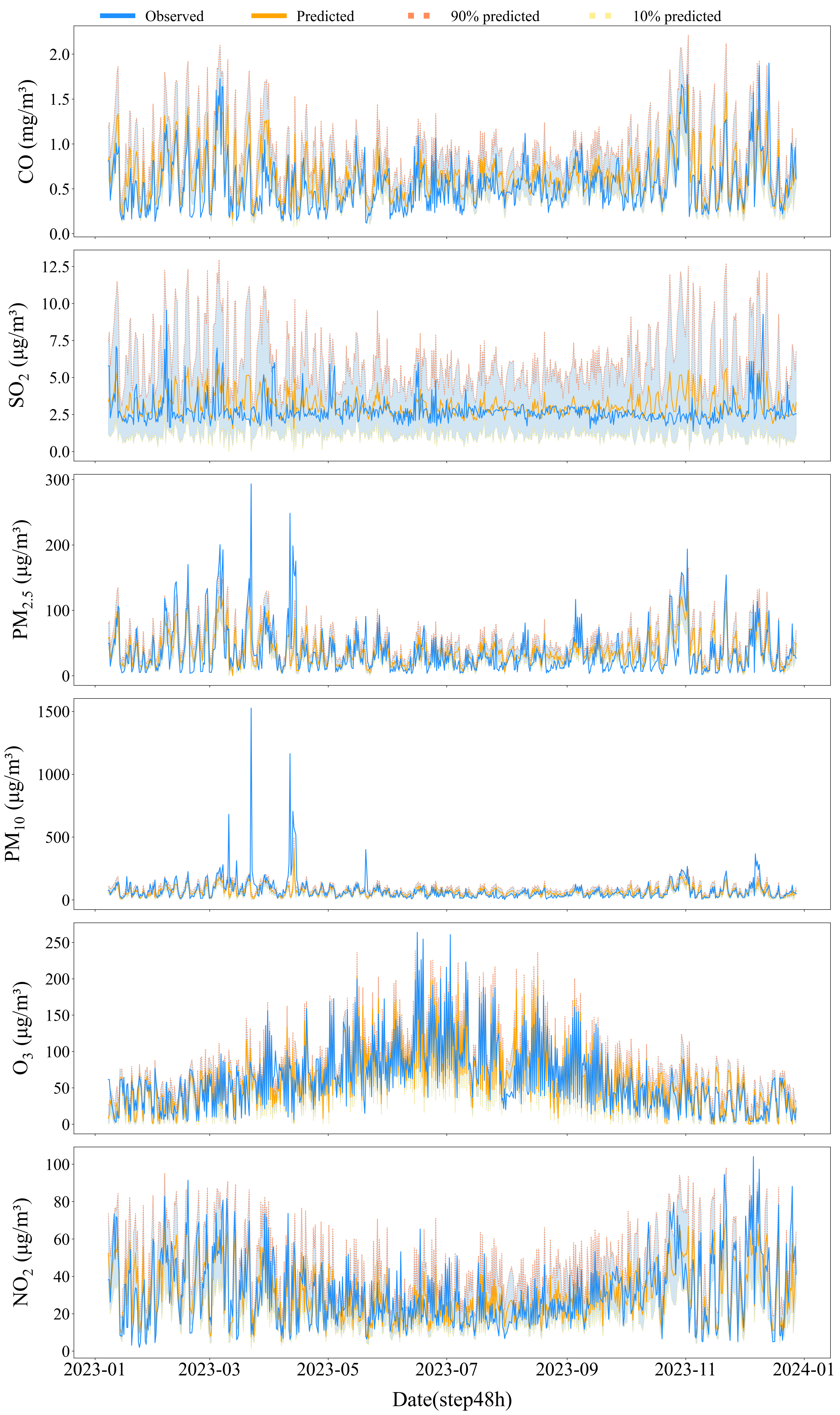}
\caption{Comparison of simulated and observed concentrations of six pollutants in beijing at 48-hour. “Forecast” refers to the point forecast, “90\% forecast” and “10\% forecast” represent the 90th and 10th percentile forecasts, respectively.}
\label{Supplementary Fig:beijing_48h}
\end{figure}

\begin{figure}[H]
\centering
\includegraphics[height=0.7\textheight]{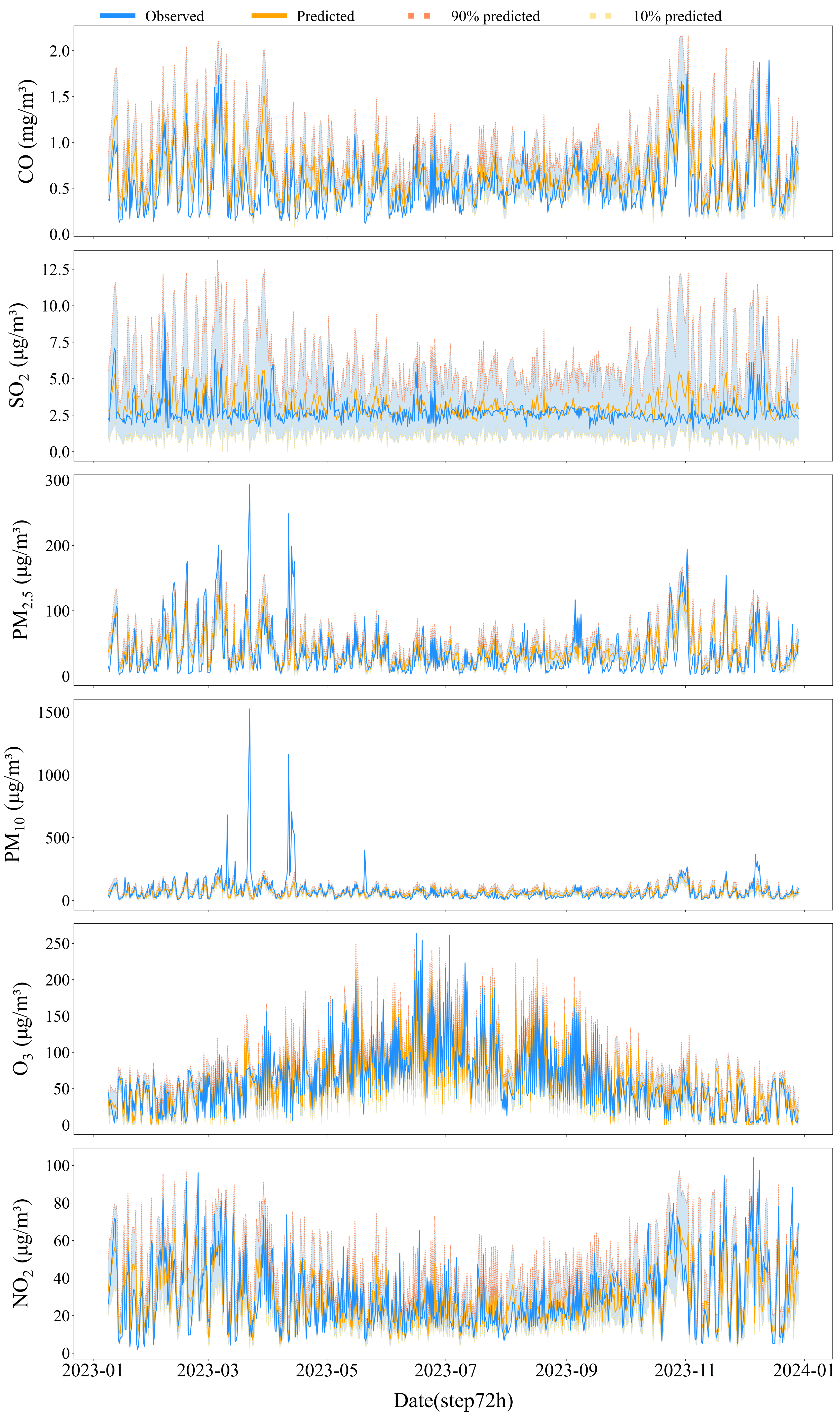}
\caption{Comparison of simulated and observed concentrations of six pollutants in beijing at 72-hour. “Forecast” refers to the point forecast, “90\% forecast” and “10\% forecast” represent the 90th and 10th percentile forecasts, respectively.}
\label{Supplementary Fig:beijing_72h}
\end{figure}

\begin{figure}[H]
\centering
\includegraphics[height=0.7\textheight]{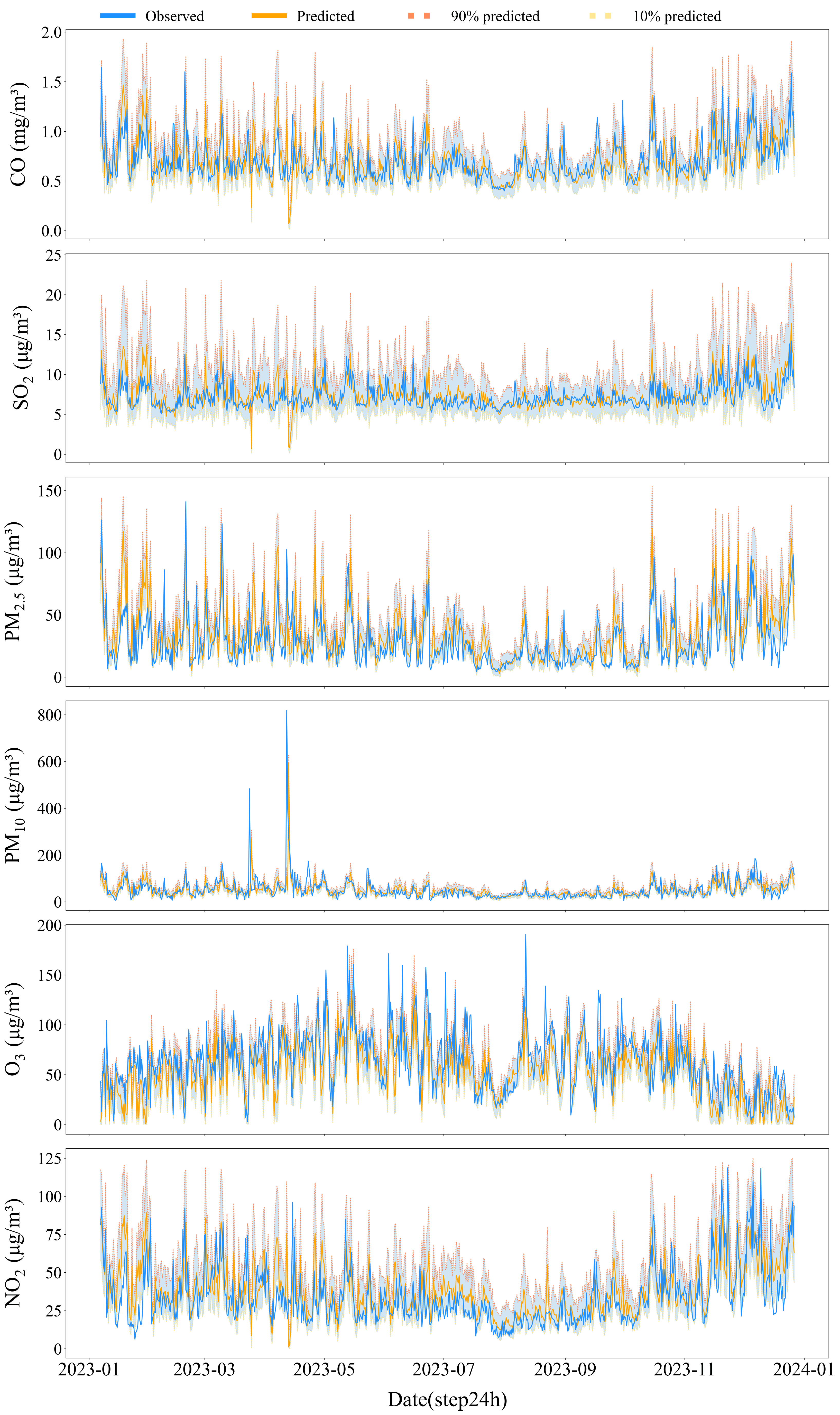}
\caption{Comparison of simulated and observed concentrations of six pollutants in shanghai at 24-hour. “Forecast” refers to the point forecast, “90\% forecast” and “10\% forecast” represent the 90th and 10th percentile forecasts, respectively.}
\label{Supplementary Fig:shanghai_24h}
\end{figure}

\begin{figure}[H]
\centering
\includegraphics[height=0.7\textheight]{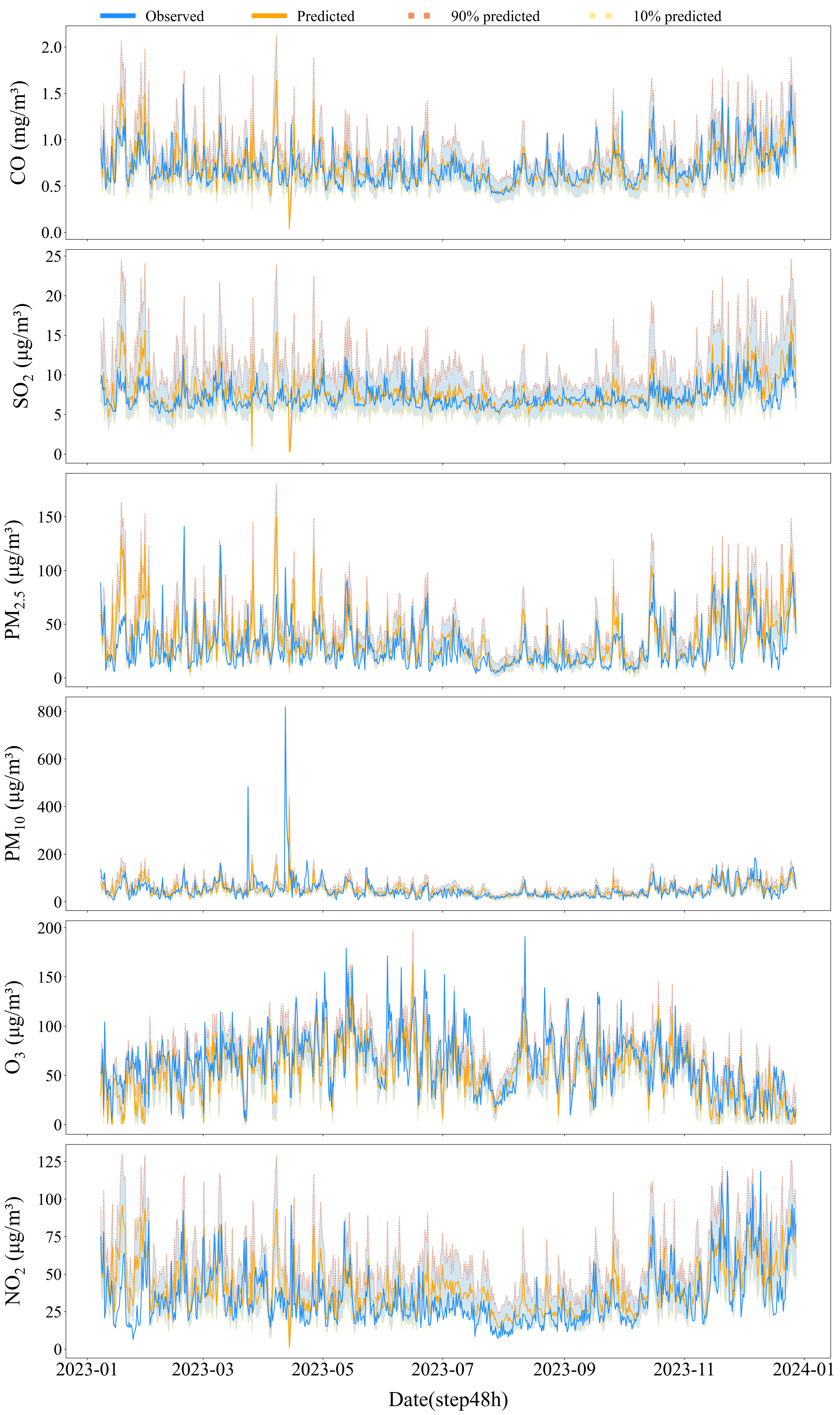}
\caption{Comparison of simulated and observed concentrations of six pollutants in shanghai at 48-hour. “Forecast” refers to the point forecast, “90\% forecast” and “10\% forecast” represent the 90th and 10th percentile forecasts, respectively.}
\label{Supplementary Fig:shanghai_48h}
\end{figure}

\begin{figure}[H]
\centering
\includegraphics[height=0.7\textheight]{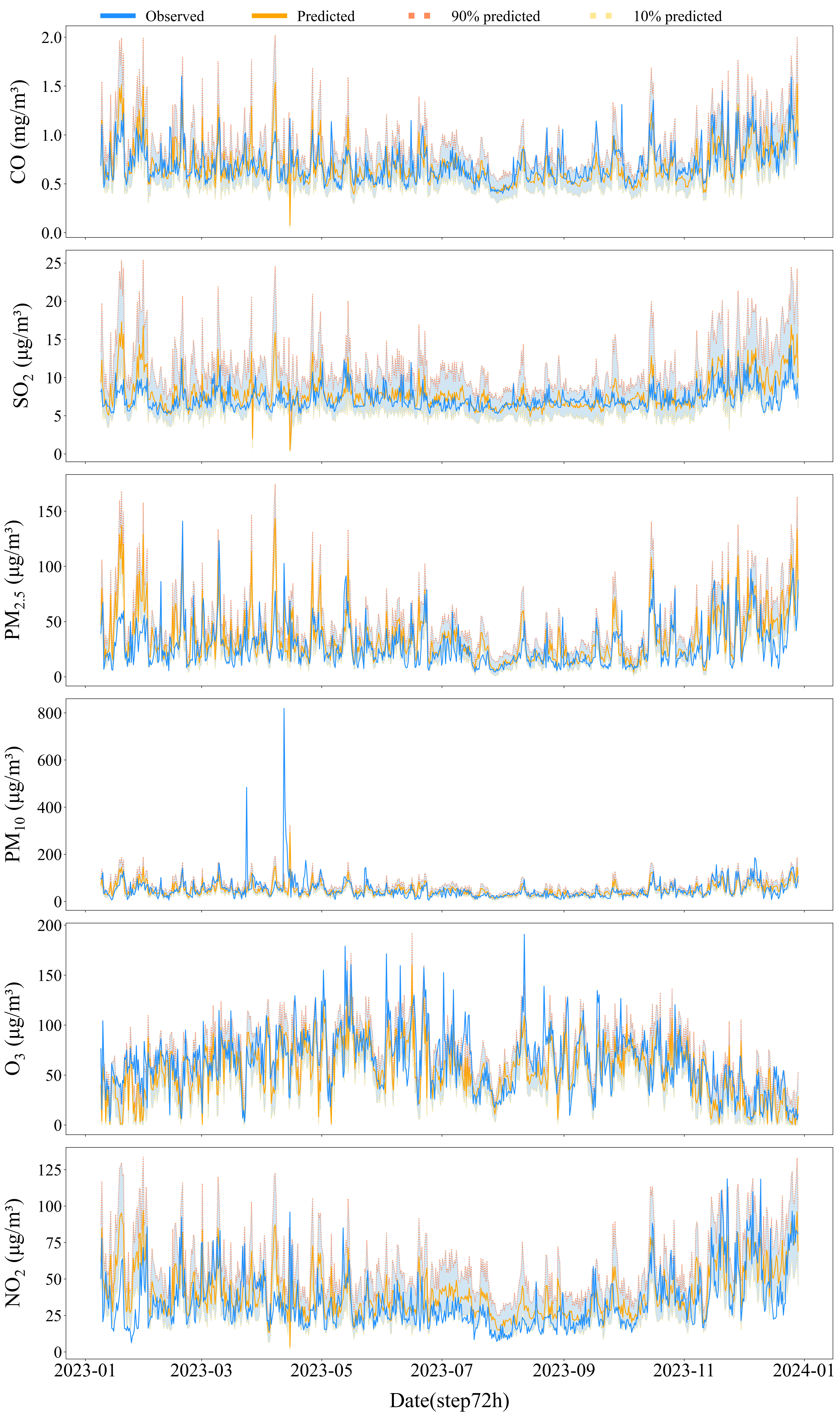}
\caption{Comparison of simulated and observed concentrations of six pollutants in shanghai at 72-hour. “Forecast” refers to the point forecast, “90\% forecast” and “10\% forecast” represent the 90th and 10th percentile forecasts, respectively.}
\label{Supplementary Fig:shanghai_72h}
\end{figure}

\begin{figure}[H]
\centering
\includegraphics[height=0.7\textheight]{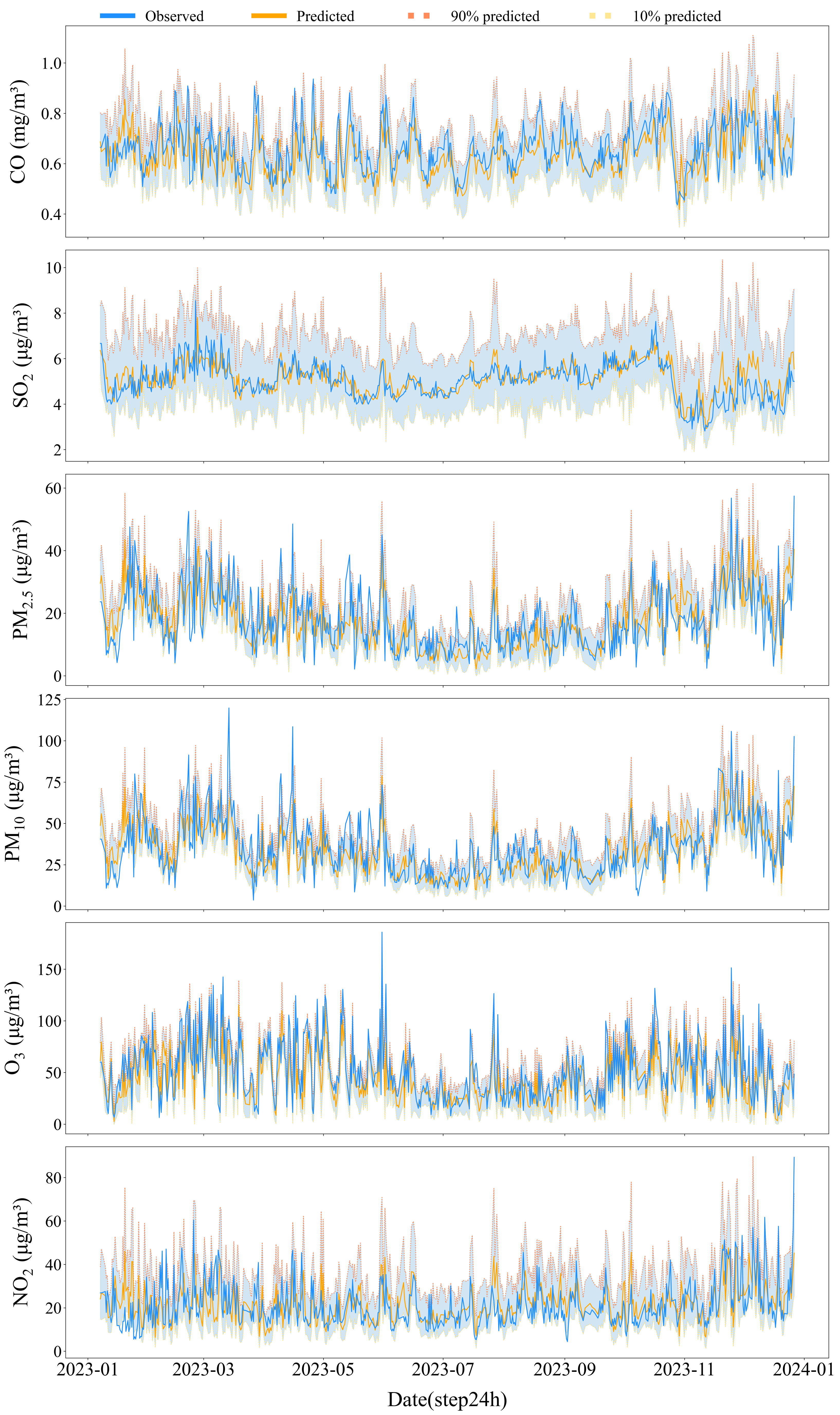}
\caption{Comparison of simulated and observed concentrations of six pollutants in shenzhen at 24-hour. “Forecast” refers to the point forecast, “90\% forecast” and “10\% forecast” represent the 90th and 10th percentile forecasts, respectively.}
\label{Supplementary Fig:shenzhen_24h}
\end{figure}

\begin{figure}[H]
\centering
\includegraphics[height=0.7\textheight]{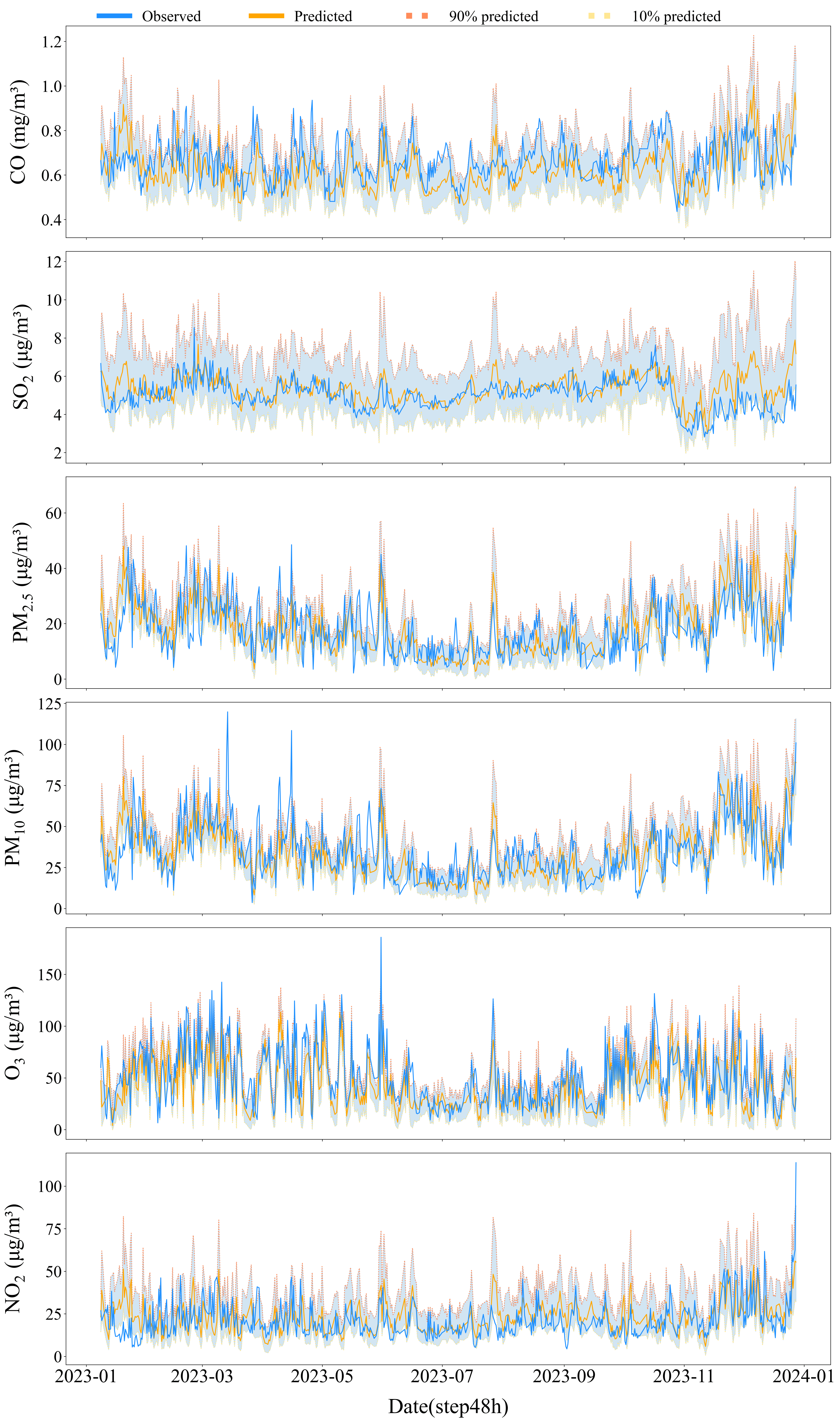}
\caption{Comparison of simulated and observed concentrations of six pollutants in shenzhen at 48-hour. “Forecast” refers to the point forecast, “90\% forecast” and “10\% forecast” represent the 90th and 10th percentile forecasts, respectively.}
\label{Supplementary Fig:shenzhen_48h}
\end{figure}

\begin{figure}[H]
\centering
\includegraphics[height=0.7\textheight]{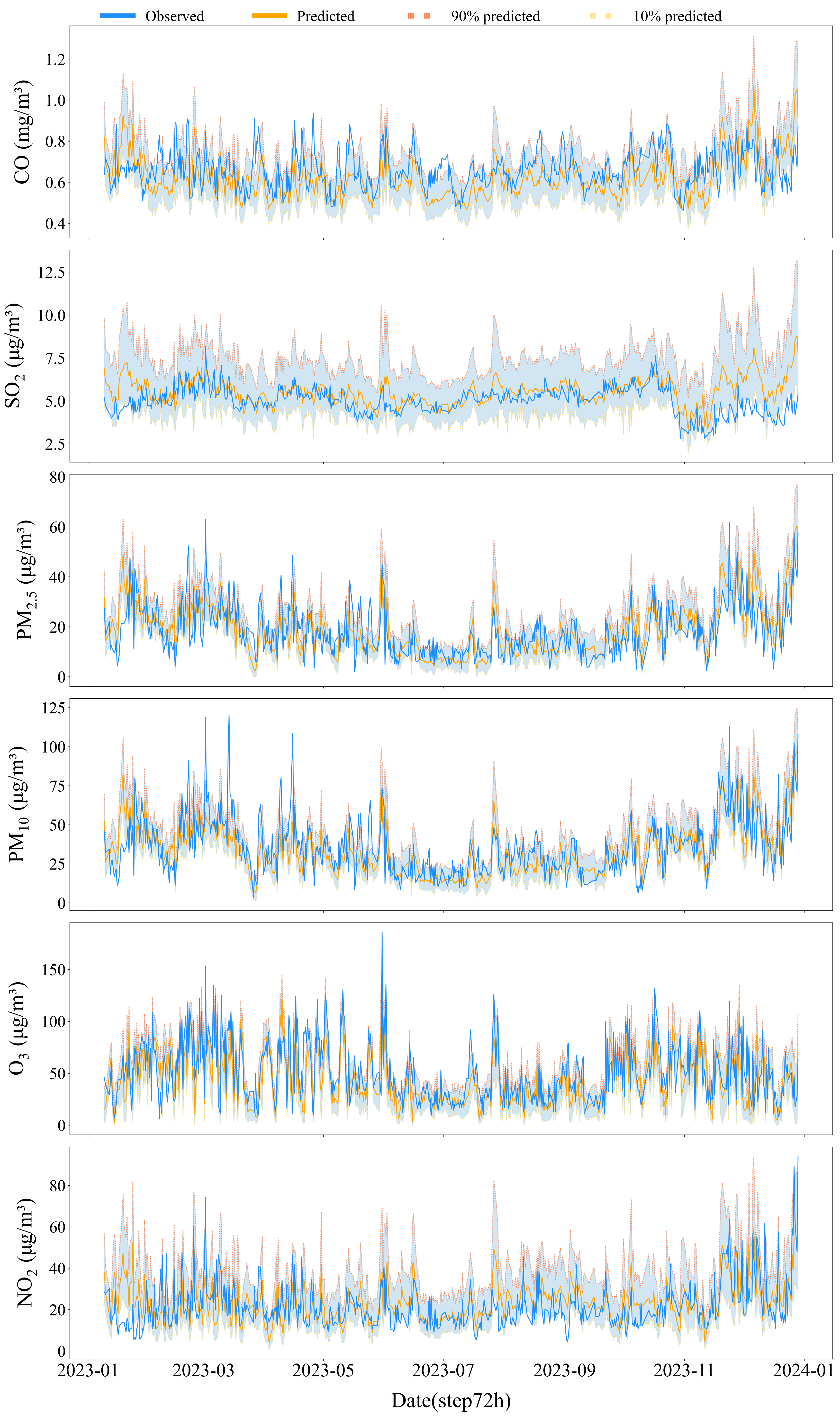}
\caption{Comparison of simulated and observed concentrations of six pollutants in shenzhen at 72-hour. “Forecast” refers to the point forecast, “90\% forecast” and “10\% forecast” represent the 90th and 10th percentile forecasts, respectively.}
\label{Supplementary Fig:shenzhen_72h}
\end{figure}

\begin{figure}[H]
\centering
\includegraphics[width=\linewidth]{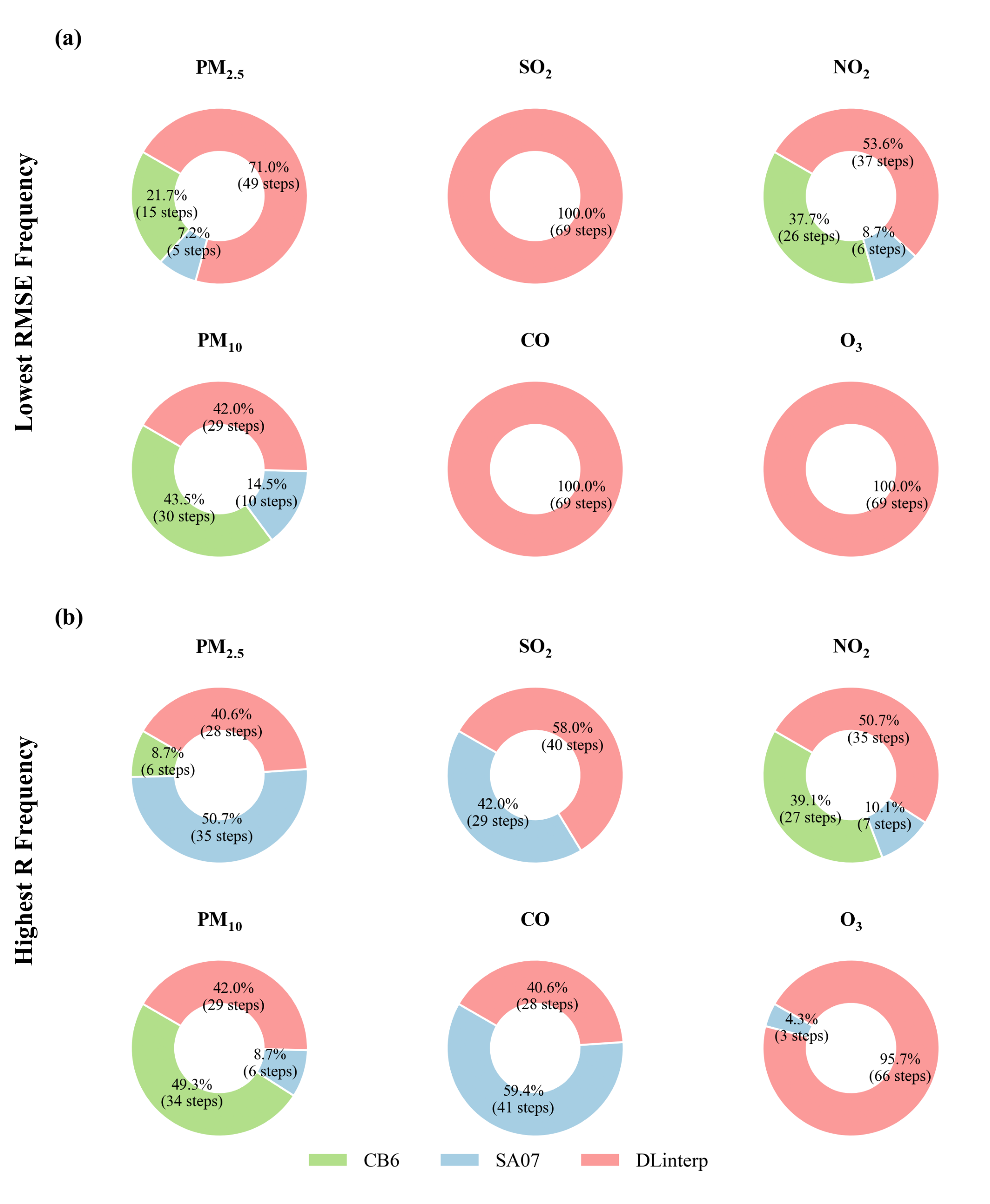}
\caption{Comparison of 72-hour forecasts between the FuXi-Air and numerical models in Shanghai.Due to differences in forecast initialization times, only steps 4 to 72 are compared here.(a) shows the proportion of time steps (out of 69) where each method achieves the lowest RMSE — higher values indicate more stable and accurate predictions. (b) shows the proportion of time steps where each method achieves the highest Pearson correlation coefficient (R) — higher values indicate better agreement with observed data.}
\label{Supplementary Fig:3.2.3FuXi-Air_vs._WRF-CMAQ}
\end{figure}

\begin{figure}[H]
\centering
\includegraphics[width=\linewidth]{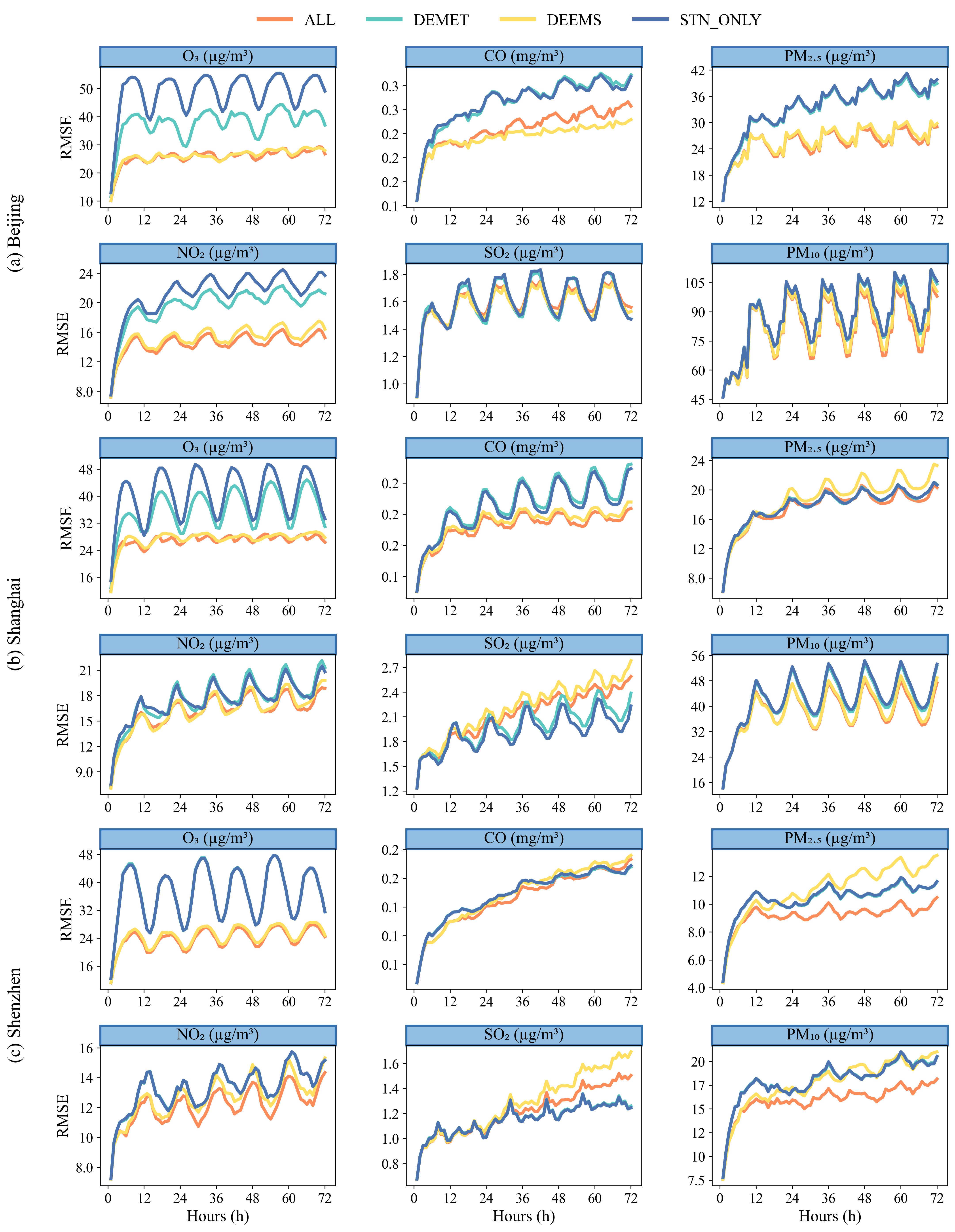}
\caption{RMSE curves from ablation experiments in Beijing, Shanghai, and Shenzhen}
\label{Supplementary Fig:rmse}
\end{figure}

\begin{table}[H]
    \centering
    \scriptsize
    \renewcommand{\arraystretch}{1.3} 
    \setlength{\tabcolsep}{1.2pt}  
    \caption{Performance evaluation results of ablation experiments in different cities, where the unit of RMSEs for CO is mg/m³, while for the other pollutants it is $\mu\text{g/m}^3$; MREs and R are unitless}
    \label{tab:Supplementary Table 3}
    
\begin{tabular}{cccccccccccc}
\hline
\multirow{2}{*}{Exp. ID}                                                 & \multicolumn{2}{c}{City}      & \multicolumn{3}{c}{Beijing} & \multicolumn{3}{c}{Shanghai} & \multicolumn{3}{c}{Shenzhen} \\ \cline{2-12} 
                                                                         & \multicolumn{2}{c}{Indicator} & R       & RMSEs     & MREs    & R       & RMSEs     & MREs    & R       & RMSEs     & MREs     \\ \hline
\multirow{6}{*}{\begin{tabular}[c]{@{}c@{}}A\\ (ALL)\end{tabular}}       & O$_{3}$             & 1-72h        & 0.86    & 25.88    & 0.31   & 0.79    & 26.87    & 0.3     & 0.78    & 24.13    & 0.3     \\
                                                                         & NO$_{2}$            & 1-72h        & 0.74    & 14.59    & 0.4    & 0.67    & 16.23    & 0.43    & 0.57    & 12.17    & 0.46    \\
                                                                         & PM$_{2.5}$          & 1-72h        & 0.73    & 25.67    & 0.48   & 0.67    & 17.8     & 0.47    & 0.63    & 9.24     & 0.41    \\
                                                                         & PM$_{10}$           & 1-72h        & 0.47    & 83.3     & 0.41   & 0.5     & 39.01    & 0.4     & 0.67    & 16.03    & 0.33    \\
                                                                         & CO             & 1-72h        & 0.7     & 0.25     & 0.39   & 0.63    & 0.19     & 0.2     & 0.57    & 0.13     & 0.15    \\
                                                                         & SO$_{2}$            & 1-72h        & 0.34    & 1.59     & 0.4    & 0.56    & 2.14     & 0.22    & 0.71    & 1.23     & 0.19    \\
\multirow{6}{*}{\begin{tabular}[c]{@{}c@{}}B\\ (DEMET)\end{tabular}}     & O$_{3}$             & 1-72h        & 0.66    & 38.22    & 0.48   & 0.5     & 36.36    & 0.42    & 0.33    & 37.72    & 0.45    \\
                                                                         & NO$_{2}$            & 1-72h        & 0.45    & 19.8     & 0.52   & 0.53    & 17.71    & 0.4     & 0.42    & 13.34    & 0.49    \\
                                                                         & PM$_{2.5}$          & 1-72h        & 0.44    & 33.88    & 0.63   & 0.49    & 18.05    & 0.45    & 0.46    & 10.47    & 0.47    \\
                                                                         & PM$_{10}$           & 1-72h        & 0.36    & 88.9     & 0.54   & 0.38    & 42.73    & 0.42    & 0.53    & 18.15    & 0.38    \\
                                                                         & CO             & 1-72h        & 0.39    & 0.3      & 0.45   & 0.47    & 0.22     & 0.22    & 0.55    & 0.14     & 0.16    \\
                                                                         & SO$_{2}$            & 1-72h        & 0.25    & 1.6      & 0.37   & 0.58    & 2.01     & 0.19    & 0.74    & 1.16     & 0.17    \\
\multirow{6}{*}{\begin{tabular}[c]{@{}c@{}}C\\ (DEEMS)\end{tabular}}     & O$_{3}$             & 1-72h        & 0.86    & 26.18    & 0.32   & 0.78    & 27.42    & 0.31    & 0.76    & 24.85    & 0.31    \\
                                                                         & NO$_{2}$            & 1-72h        & 0.72    & 15.31    & 0.41   & 0.69    & 16.36    & 0.43    & 0.54    & 12.77    & 0.51    \\
                                                                         & PM$_{2.5}$          & 1-72h        & 0.71    & 26.18    & 0.46   & 0.66    & 19.03    & 0.51    & 0.58    & 11.19    & 0.48    \\
                                                                         & PM$_{10}$           & 1-72h        & 0.46    & 84.8     & 0.42   & 0.5     & 39.44    & 0.42    & 0.63    & 17.93    & 0.38    \\
                                                                         & CO             & 1-72h        & 0.69    & 0.24     & 0.36   & 0.63    & 0.19     & 0.21    & 0.55    & 0.14     & 0.16    \\
                                                                         & SO$_{2}$            & 1-72h        & 0.35    & 1.58     & 0.39   & 0.54    & 2.23     & 0.23    & 0.69    & 1.31     & 0.2     \\
\multirow{6}{*}{\begin{tabular}[c]{@{}c@{}}D\\ (STN\_ONLY)\end{tabular}} & O$_{3}$             & 1-72h        & 0.22    & 49.26    & 0.64   & 0.27    & 41.01    & 0.46    & 0.33    & 37.66    & 0.45    \\
                                                                         & NO$_{2}$            & 1-72h        & 0.33    & 21.45    & 0.54   & 0.51    & 17.62    & 0.42    & 0.42    & 13.35    & 0.49    \\
                                                                         & PM$_{2.5}$          & 1-72h        & 0.42    & 34.23    & 0.63   & 0.47    & 18.06    & 0.49    & 0.46    & 10.51    & 0.48    \\
                                                                         & PM$_{10}$           & 1-72h        & 0.35    & 89.55    & 0.53   & 0.36    & 43.33    & 0.43    & 0.53    & 18.15    & 0.37    \\
                                                                         & CO             & 1-72h        & 0.4     & 0.3      & 0.45   & 0.44    & 0.21     & 0.21    & 0.55    & 0.14     & 0.16    \\
                                                                         & SO$_{2}$            & 1-72h        & 0.23    & 1.61     & 0.37   & 0.59    & 1.93     & 0.18    & 0.75    & 1.16     & 0.17    \\ \hline
\end{tabular}
\end{table}

\end{document}